\documentclass{article}

\usepackage{PRIMEarxiv}

\usepackage[utf8]{inputenc} % allow utf-8 input
\usepackage[T1]{fontenc}    % use 8-bit T1 fonts
\usepackage{hyperref}       % hyperlinks
\usepackage{url}            % simple URL typesetting
\usepackage{booktabs}       % professional-quality tables
\usepackage{amsfonts}       % blackboard math symbols
\usepackage{nicefrac}       % compact symbols for 1/2, etc.
\usepackage{microtype}      % microtypography
\usepackage{lipsum}
\usepackage{fancyhdr}       % header
\usepackage{graphicx}       % graphics
\graphicspath{{media/}}     % organize your images and other figures under media/ folder

\usepackage{amsmath}
\usepackage{booktabs}
\usepackage{graphicx}
\usepackage{subcaption}
\usepackage{algorithm}
\usepackage{algorithmic}
\usepackage{xcolor}
\usepackage{bm}
\usepackage{amssymb}

\title{EIA-SEC: Improved Actor-Critic Framework for Multi-UAV Collaborative Control in Smart Agriculture}

\author{Quanxi~Zhou$^{1}$\thanks{The first two authors contributed equally to this work.},
Wencan~Mao$^{2*}$, Yilei~Liang$^{3}$,~Manabu~Tsukada$^{1}$,~Yunling Liu$^{4}$\thanks{The corresponding auther is Yunling Liu (yunling@cau.edu.cn).},~Jon Crowcroft~$^{3}$\\
$^{1}$The University of Tokyo, Tokyo, Japan\\
$^{2}$National Institute of Informatics, Tokyo, Japan\\
$^{3}$The University of Cambridge, Cambridge, UK \\
$^{4}$China Agricultural University, Beijing, China \\
\{usainzhou, mtsukada\}@g.ecc.u-tokyo.ac.jp, wencan\_mao@nii.ac.jp, yl841@cam.ac.uk,\\ jon.crowcraft@cl.cam.ac.uk, yunling@cau.edu.cn.
}

\begin{document}
\maketitle
\begin{abstract}
The widespread application of wireless communication technology has promoted the development of smart agriculture, where unmanned aerial vehicles (UAVs) play a multifunctional role.
We target a multi-UAV smart agriculture system where UAVs cooperatively perform data collection, image acquisition, and communication tasks.
In this context, we model a Markov decision process to solve the multi-UAV trajectory planning problem.
Moreover, we propose a novel Elite Imitation Actor-Shared Ensemble Critic (EIA-SEC) framework, where agents adaptively learn from the elite agent to reduce trial-and-error costs, and a shared ensemble critic collaborates with each agent’s local critic to ensure unbiased objective value estimates and prevent overestimation. 
Experimental results demonstrate that EIA-SEC outperforms state-of-the-art baselines in terms of reward performance, training stability, and convergence speed.
\end{abstract}

% keywords can be removed

\section{Introduction}
\label{introduction}

The proliferation of digitization technology~\cite{digital} and the fifth generation of cellular network technology (5G)~\cite{5g} has boosted the development of smart agriculture~\cite{smart-agri}.
Instead of manually monitoring/analyzing farmland, wireless sensor networks (WSNs) have enabled widespread applications of agricultural data acquisition~\cite{a-data}, intelligent monitoring~\cite{a-monitoring}, precision agriculture analytics~\cite{precision}, and the implementation of the Agricultural Internet of Things (Agri-IoT)~\cite{a-IoT}.

Meanwhile, the development of unmanned aerial vehicle (UAV) technology has significantly advanced intelligent agriculture~\cite{agri-uav}. On one hand, UAVs can collect data from WSNs, enabling real-time monitoring of crop growth, efficient data aggregation, and timely environmental analysis~\cite{agri-link}. On the other hand, UAVs equipped with cameras can continuously capture aerial imagery of farmland, supporting visual analysis for crop health assessment, pest detection, and disaster monitoring~\cite{uav-image-agri}. 
Moreover, in remote and rugged areas, cellular communication might become inaccessible or be obstructed by obstacles.
To address this issue, some UAVs can assist with aerial communication, ensuring reliable information exchange, location sharing, and data aggregation among UAVs without deploying additional ground stations or incurring extra infrastructure costs~\cite{uav-hoc}.

In a multi-UAV smart agricultural system, where the UAVs responsible for data collection, image acquisition, and communication co-exist, different types of UAVs are supposed to collaborate with each other while addressing their respective task requirements.
Despite heuristics~\cite{heuristic}, deep reinforcement learning (DRL)~\cite{DRL}, hybrid optimization~\cite{hybrid}, and convex optimization~\cite{convex} have been widely used for trajectory planning for UAVs, there lacks a unified framework for multi-UAV collaborative control in smart agriculture. 

However, there are challenges associated with this problem.
From a control perspective, each UAV must avoid collision with others~\cite{collision}. In addition, limited battery capacity and UAV dynamics must be taken into account when making joint decisions, requiring UAVs to adopt energy-efficient strategies~\cite{battery}.
From a communication perspective, UAVs must consider the distance between each other and WSNs to ensure effective information exchange and reliable data transmission~\cite{air-to-air}.
From an application perspective, the quality of information (QoI) metrics include the age of information (AoI) of WSN data collection and visual data freshness (VDF) of image acquisition.
If not well planned, competition between UAVs ~\cite{competition} will lead to a reduction in task efficiency and overall QoI performance.

To address the above issues, we propose \textit{a novel elite imitation actor-shared ensemble critic (EIA-SEC)} framework, where each agent is a UAV.
To enable agents to learn from high-performing peers with the same type of tasks, we periodically select an elite during training, and allow the actor networks of other agents to softly learn from the actor parameters of the elite agent. This approach reduces trial-and-error costs and guides agents more effectively towards optimal decision directions. As training progresses, agents gradually decrease their reliance on the elite, thereby maintaining exploration and preventing convergence to local optima. In this way, this elite imitation actor (EIA) mechanism improves both the reward performance and the convergence speed of the algorithm, enabling agents to learn effectively and efficiently.

In addition, to ensure the unbiased evaluation of action-state values for each type of task, we equip each agent not only with a local critic, but also with shared ensemble critic (SEC) networks for agents performing the same type of task. During each value evaluation, the local agent critic collaborates with the SEC, contributing to the action-value proportionally. Furthermore, whenever a local critic is updated, the SEC is synchronized accordingly. This mechanism effectively prevents overestimation of Q-values by individual critics, thereby enhancing the stability and convergence of the algorithm.
Our key contributions are as follows:
\begin{itemize}
    \item We target a multi-UAV smart agricultural system, where the UAVs responsible for data collection, image acquisition, and communication tasks co-exist. We construct a Markov decision process (MDP) to tackle the multi-UAV trajectory planning problem for various tasks, considering factors such as AoI, VDF, energy consumption, collision avoidance, and communication quality.
    
    \item We propose a \textit{novel EIA-SEC} framework that enhances learning efficiency and stability in multi-agent systems. Agents periodically elect elite individuals, and other agents adaptively learn from their actor parameters to reduce trial-and-error costs. In addition, a shared ensemble critic collaborates with each agent’s local critic to ensure unbiased objective value estimates and prevent overestimation. This combination accelerates convergence, improves performance, and maintains stability in complex collaborative environments.

\end{itemize}

\section{Related Works}
\label{related_works}
% Intelligent Agriculture
UAVs have played increasingly critical roles in intelligent agriculture~\cite{uav-smart-agri}, which range from data collection~\cite{uav-data-iot, uav-aoi-iot}, visual analysis and monitoring~\cite{agri-visual}, and communication~\cite{comm-uav-agri}.
While previous studies focused on a single role, our work considers a multifunctional smart agriculture framework integrating the above functionalities.
% UAVs have played an increasingly critical role in intelligent agriculture~\cite{uav-smart-agri}. In~\cite{uav-data-iot}, researchers constructed a UAV-assisted intelligent agricultural Internet of Things (IoT) system, in which UAVs collect data generated by the WSNs. In~\cite{uav-aoi-iot}, researchers utilized the multi-UAV collaborative system to reduce the system AoI of the intelligent IoTs. In~\cite{agri-visual}, an end-to-end observing system based on UAV to support precision agriculture and the forest monitoring has been designed.
% In~\cite{comm-uav-agri}, researchers considered the UAV communication condition in a UAV-WSN intelligent agricultural system. While these previous studies have explored UAV-assisted smart agriculture only focus on isolated intelligent function. This work aims to address these limitations by integrating their strengths into a comprehensive multifunctional smart agriculture framework, incorporating data sensing and collection, UAV-based monitoring, and UAV-assisted communication.
Furthermore, trajectory planning is a vital research topic for the multi-UAV collaborative system.
Traditional methods of UAV trajectory planning include heuristic methods~\cite{ACO-A}, hybrid optimization techniques~\cite{hybrid_pso}, and convex optimization~\cite{convex_op}. 
% For instance, \cite{ACO-A} proposed a combination of Ant Colony Optimization (ACO) and the A* algorithm to optimize UAV trajectories. A hybrid Particle Swarm Optimization (PSO) approach was introduced in \cite{hybrid_pso} to enhance task performance in UAV path planning. Convex optimization has also been applied to linearize the dynamic model challenges for UAV trajectory design~\cite{convex_op}.
Despite their effectiveness in controlled settings, these methods face substantial limitations in real-world applications, such as fixed parameters, high computational complexity, and low adaptation to environmental uncertainty. 
% Firstly, they typically assume fixed parameters, such as predefined UAV starting points and destinations, which limits their ability to generalize and adapt policies dynamically in changing environments. Secondly, the high computational complexity of these algorithms often renders them unsuitable for scenarios with stringent real-time requirements. Finally, these algorithms struggle to account for the interactions among multiple heterogeneous agents in a scenario, each with different objectives and fitness rewards. Consequently, they face significant challenges in trajectory optimization for multiple UAVs performing diverse tasks.

\begin{figure*}[t]
    \centering
    \includegraphics[width=0.85\textwidth, keepaspectratio]{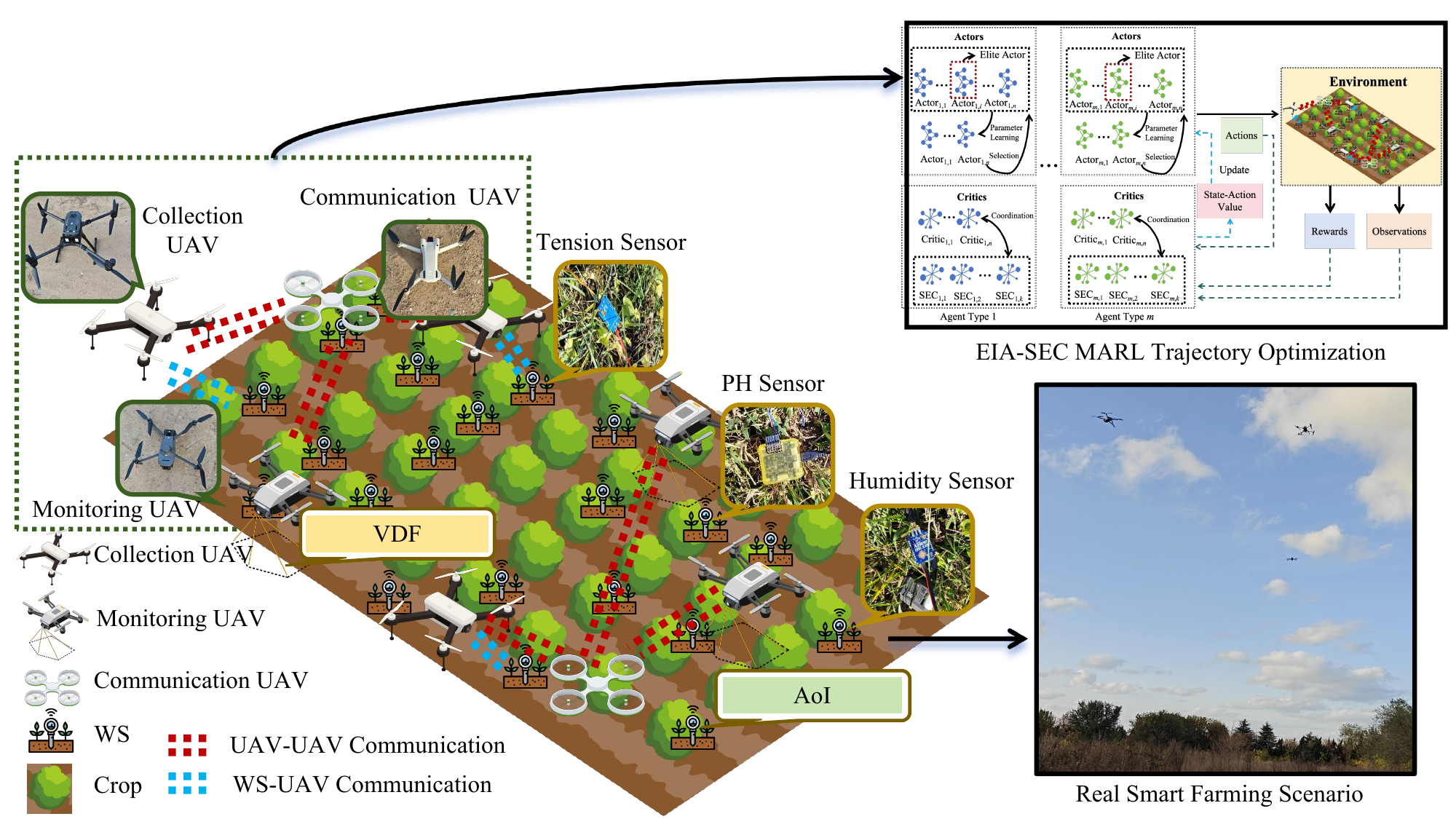}
    \caption{The envisioned multi-UAV collaborative system for smart agriculture, where collection UAVs, monitoring UAVs, and communication UAVs cooperate with each other to minimize the system-level AoI and VDF.}
    \label{fig: overview}
    \setlength{\belowcaptionskip}{-10pt}
\end{figure*}

% DRL Trajectory Planning
Compared with the above algorithms, deep reinforcement learning (DRL) excels in multi-UAV decision-making scenarios. 
% offers several key advantages, including adaptability to dynamic environments, the ability to handle high-dimensional tasks, support for online learning, and relatively low computational complexity in multi-UAV decision-making scenarios. 
DRL methods, such as deep deterministic policy gradient (DDPG)~\cite{ddpg}, soft actor-critic (SAC)~\cite{sac}, twin delayed DDPG (TD3)~\cite{td3}, proximal policy optimization (PPO)~\cite{ppo}, and counterfactual multi-agent policy gradients (COMA)~\cite{COMA_real}, have been widely used.
% For example, \cite{ddpg} highlighted that continuous-action DRL algorithms, such as Deep Deterministic Policy Gradient (DDPG), achieve faster decision-making and lower computational complexity than other DRL approaches in UAV trajectory planning. In \cite{sac}, the soft actor-critic (SAC) algorithm was employed to maximize the sum rate, demonstrating superior performance over deep Q-learning (DQN), which is limited to discrete action spaces. Twin Delayed DDPG (TD3) was applied in \cite{td3} for UAV trajectory planning, with experiments showing strong convergence and robust task performance. Furthermore, PPO was applied for UAV obstacle avoidance in \cite{ppo}, showing feasibility and generalizability; however, but it suffers from low sample efficiency, susceptibility to local optima, and limited multi-agent coordination, which can hinder performance in multi-UAV scenarios.
Furthermore, various modifications and enhancements of DRLs have emerged in recent years.
In~\cite{REDQ}, Randomized Ensemble Double Q-learning (REDQ) was proposed to improve training stability and data efficiency by performing random sampling during updates. However, REDQ relies on extensive environment interactions and requires training and maintaining multiple Q networks, resulting in high computational and memory overhead. 
% In~\cite{COMA}, a counterfactual multi-agent (COMA) policy gradient was proposed for centralized critic evaluation. However, COMA suffers from high computational complexity and relies on centralized training, limiting its scalability. % As agent numbers grow, computing the counterfactual baseline and handling the joint action space become increasingly difficult, restricting its use in large-scale multi-agent tasks. 
In~\cite{el}, an ensemble learning bootstrapped DQN was designed to leverage the ensemble approach to reduce variance and improve learning stability, but it struggles with the continuous action space tasks and multi-functional agent collaborative systems.
% To address the challenges of trajectory optimization in heterogeneous multi-agent systems with diverse functions, 
In this work, we propose a novel EIA-SEC framework, which enhances learning efficiency and stability in complex, dynamic, and cooperative multi-agent environments.
% In this framework, agents periodically elect elite individuals, while other agents adaptively learn from the elite actors’ parameters, reducing trial-and-error costs. A shared ensemble critic collaborates with each agent’s local critic to provide reliable objective value estimates and prevent overestimation, thereby mitigating variance and instability issues common in single-critic approaches. This design accelerates convergence, improves overall performance, and maintains stability in challenging multi-agent scenarios.

\section{System Overview}
\label{system_overview}
% This section demonstrates an exemplary scenario and introduces the employed methodology.
% \subsection{Exemplary Scenario}

Fig.~\ref{fig: overview} demonstrates 
% an exemplary scenario located in an intelligent farmland, where 
a collaborative system consisting of $n_{\text{c}}$ communication $\text{UAV}_{\text{c}}$, $n_{\text{m}}$ monitoring $\text{UAV}_{\text{m}}$, and $n_{\text{d}}$ data collection $\text{UAV}_{\text{d}}$ to perform smart agricultural tasks.
% We define communication $ \text{UAV}_{\text{c}} =\{ \text{UAV}_{\text{c},1}, \text{UAV}_{\text{c},2},\dots,\text{UAV}_{\text{c},n_{\text{c}}}  \}$. 
% Also, we define monitoring $\text{UAV}_{\text{m}} = \{ \text{UAV}_{\text{m},1}, \text{UAV}_{\text{m},2},\dots,\text{UAV}_{\text{m},n_{\text{m}}}\}$, and data collection $\text{UAV}_{\text{d}} = \{ \text{UAV}_{\text{d},1}, \text{UAV}_{\text{d},2},\dots,\text{UAV}_{\text{d},n_{\text{d}}}  \}$.  
% Assume that these UAVs perform their own tasks at \textcolor{red}{heights $H_{\text{UAV}_{\text{c}}}, H_{\text{UAV}_{\text{m}}}, H_{\text{UAV}_{\text{d}}}$, respectively}.
The position of the $i$-th communication UAV at time $t$ is expressed as 
$\bm{U}_\text{c}(i,t) = \{ U_{\text{c}x}(i,t), U_{\text{c}y}(i,t),H_{\text{UAV}} \}$, 
where each element represents the coordinates along the $x$-, $y$-, and $z$-axes in the 3D spatial coordinate system, so as the monitoring and data collection UAVs.
% Similarly, the positions of the $i$-th monitoring and data collection UAVs are denoted as 
% $\bm{U}_\text{m}(i,t) = \{ U_{\text{m}x}(i,t), U_{\text{m}y}(i,t), H_{\text{UAV}} \}$ and 
% $\bm{U}_\text{d}(i,t) = \{ U_{\text{d}x}(i,t), U_{\text{d}y}(i,t), H_{\text{UAV}} \}$, respectively.
For each UAV, the 2D task space can be defined as $\mathbb{R}^{2}:\{x\in[U_{x-\text{min}},U_{x-\text{max}}],y\in [U_{y-\text{min}}, U_{y-\text{max}}],H_{\text{c}}|H_{\text{m}}|H_{\text{d}}\}$, where $H_{\text{c}}$, $H_{\text{m}}$, and $H_{\text{d}}$ represent the flight height of communication $\text{UAV}_{\text{c}}$, monitoring $\text{UAV}_{\text{m}}$, and data collection $\text{UAV}_{\text{d}}$, respectively.
The maximum velocities of the communication, monitoring, and data collection UAVs 
are assumed to be identical, denoted as $V_{x\text{-}\max}$ and $V_{y\text{-}\max}$ 
along the $x$ and $y$ directions, respectively.

The data collection UAVs are responsible for gathering information acquired by the WSN, which consists of multiple categories of wireless sensors (WSs) that are irregularly distributed across the farmland with certain spacing constraints, represented as $\text{WS}=\{ \text{WS}_1,\text{WS}_2,...,\text{WS}_{n_{\text{WS}}}\}$. 
In different farmland scenarios, the distribution of WSs varies with local conditions.
Each WS periodically measures soil parameters such as temperature, humidity, and pH level. 
% \textcolor{red}{ Due to the complex agricultural environment, where WSs are sparsely distributed and often obstructed by crops, % ground base stations find it difficult to communicate with the WSs directly. Therefore, 
% we utilize the multi-UAV collaborative system for data collection. 
% The WSs follow a wake-up or beacon-initiated protocol. The sensor remains in deep sleep. It only transmits data when it detects a beacon signal from the approaching UAVs.}
WSs of different types are configured with distinct data collection and transmission periods depending on their sensing tasks.
The freshness of the data is represented by the data $\text{AoI}$, which accumulates over time until out-of-date (i.e., the upper threshold $\text{AoI}_{\max}$ is reached).

To obtain real-time images of crop growth and to prevent the intrusion of pests, animals, and birds, the farmland is divided into $n_{px} \times n_{py}$ small plots, which can be represented as $\mathcal{P} = {\mathcal{P}}_{1,1},\dots,\mathcal{P}_{n_{\text{px},{\text{py}}}}$. 
The monitoring UAVs are responsible for patrolling these plots and capturing image data for each plot area. The freshness of the image information is represented by the $\text{VDF}$, which accumulates over time until out-of-date (i.e., the upper threshold $\text{VDF}_{\max}$ is reached).

The communication UAVs are responsible for providing wireless connectivity to both data collection and monitoring UAVs. 
At each time $t$, the communication UAVs allocate communication resources to the collaborative multi-UAV system based on the positions of other UAVs and themselves, using the Hungarian algorithm~\cite{hungarian}. 
The communication UAVs maintain real-time inter-UAV communication to aggregate data and positional information received from other UAVs. 
They perform data analysis and early warning computations. 
When abnormal situations are detected, the communication UAVs immediately forward the corresponding information to the ground data center (GDC) for further investigations.
Meanwhile, they
broadcast updated state information, including the status of other UAVs, the system-level AoI, and VDF, to facilitate policy decision-making for other UAVs.

\section{System Model}
\label{system_model}
% This section illustrates the communication, energy consumption, data collection, and monitoring models, and then formulates the trajectory planning problems for communication, monitoring, and collection UAVs.

\subsection{UAV Communication Model}
In the considered scenario, communication UAVs provide communication service for monitoring and collection UAVs. The service protocol can be represented as:
{%
\makeatletter
\renewcommand{\ALG@name}{Protocol} % 局部修改
\makeatother
\begin{algorithm}
 \caption{The communication service protocol}
 \begin{algorithmic}[1]
   \STATE $\text{UAV}_{\text{m}}$ and $\text{UAV}_{\text{d}}$ broadcast collected data and positions.
\STATE $\text{UAV}_{\text{c}}$ receives positions and data from $\text{UAV}_{\text{m}}$ and $\text{UAV}_{\text{d}}$.
\STATE $\text{UAV}_{\text{c}}$ periodically transmits the aggregated data to GDC.
\STATE $\text{UAV}_{\text{c}}$ determines optimal communication assignments using the Hungarian algorithm as follows:
\begin{itemize}
    \item Construct a cost matrix based on link quality and distance between $\text{UAV}_{\text{c}}$ and each $\text{UAV}_{\text{m}}$, $\text{UAV}_{\text{d}}$.
    \item Find the minimum-cost matching, ensuring efficient and stable connections.
    \item Allocate connections according to the optimal assignment results.
\end{itemize}
\STATE $\text{UAV}_{\text{c}}$ adjusts its position dynamically to maintain high-quality communication links.   
 \end{algorithmic}
 \label{protocol:communication}
\end{algorithm}
}%

To improve the service quality of $\text{UAV}_{\text{c}}$, we design a parameter $D_{\text{c}}(i,t)$ to measure the service performance of $\text{UAV}_{\text{c},i}$ at time $t$.

\begin{equation}
    D_{\text{c}}(i,t) = \epsilon_1 D_{\text{cg}}(i,t) + \epsilon_2 D_{\text{cf}}(i,t)+\epsilon_3 D_{\text{cc}}(i,t),
\end{equation}
where $D_{\text{cg}}(i,t)$ represents the distance between 
$\text{UAV}_{\text{c},i}$ and the geometric centroid of served UAVs, $D_{\text{cf}}(i,t)$ represents the distance between $\text{UAV}_{\text{c},i}$ and the farthest served UAV,  $D_{\text{cc}}(i,t)$ represents the distance between 
$\text{UAV}_{\text{c},i}$ and the geometric centroid of all the communication $\text{UAV}_{\text{c}}$, $\epsilon_1$, $\epsilon_2$ and $\epsilon_3$ represent the weight coefficients.

\subsection{Energy Consumption Model}
$\text{UAV}_{\text{c}}$, $\text{UAV}_{\text{m}}$, and $\text{UAV}_{\text{d}}$ follow distinct energy consumption models, reflecting their varying operational characteristics. 

As for the communication $\text{UAV}_{\text{c},i}$, we set the energy consumption $E_\text{c}(i,t)$ as:
\begin{equation}
E_\text{c}(i,t)= E_{\text{fly}}(i,t) + E_{\text{comm}}(i,t) + E_{\text{cmp}}(i,t),
\end{equation}
where $E_{\text{fly}}(i,t)$, $ E_{\text{cmp}}(i,t)$, and $E_{\text{comm}}(i,t) $ represent cumulative flight, communication, and computation energy consumption of $\text{UAV}_i$, respectively.

The cumulative flight energy consumption $E_{\text{fly}}(i,t)$ of UAV $i$ at time $t$ can be represented as:
\begin{equation}
E_{\text{fly}}(i,t) = \int_{0}^{t} P_{\text{fly}}(i,t) \mathrm{d}t,
\end{equation}
where $P_{\text{fly}}(i,t)$ denotes the instantaneous flight power of $\text{UAV}_i$ at time $t$. The flight power depends on the UAV’s aerodynamic structure and velocity, and is expressed as:
\begin{equation}
\begin{aligned} & 
P_{\text{fly}}(i, t) 
=\left\{\begin{array}{ll}
\frac{m_{\text{UAV}} g^{3 / 2}}{\sqrt{2 \rho_{\text{air}} A_{\text{UAV}}} \eta,} & |\bm{v}_{\text{c}}(i, t)|<v_{\text{th}}, \\
c_{1}|\bm{v}_{\text{c}}(i, t)|^{2}+ \frac{c_{2}}{|\bm{v}_{\text{c}}(i, t)|}\\+m_{\text{UAV}} g|\bm{v}_{\text{c}}(i, t)|, & |\bm{v}_{\text{c}}(i, t)| \geq v_{\text{th}},
\end{array}\right.
\end{aligned}
\end{equation}
where $m_{\text{UAV}}$ is the UAV mass, $g$ represents the gravitational acceleration, $\rho_{\text{air}}$ represents the air density, $A_{\text{UAV}}$ represents the frontal area of the UAV, and $\eta$ denotes the mechanical efficiency. The terms $v_x(i,t)$ and $v_y(i,t)$ correspond to the UAV velocity components along the $x$ and $y$ directions, respectively. A velocity threshold $v_{\text{th}}$ is introduced to distinguish between hovering and forward flight, as these states exhibit significantly different power consumption behaviors.

The aerodynamic coefficients $c_1$, $c_2$, and the frontal area $A_{\text{UAV}}$ can be defined as:
\begin{equation}
c_1 = \frac{1}{2}\rho_{\text{air}}A_{\text{UAV}}C_d,
\end{equation}
\begin{equation}
c_2 = \frac{m_{\text{UAV}}^{2}}{\eta\rho_{\text{air}}n_{\text{prp}}\pi R_{\text{prp}}^{2}},
\end{equation}
\begin{equation}
A_{\text{UAV}} = A_{\text{surf}} + n_{\text{prp}}\pi R_{\text{prp}}^{2},
\end{equation}
where $C_d$ denotes the air viscosity coefficient, $n_{\text{prp}}$ represents the number of propellers, $R_{\text{prp}}$ represents the radius of each propeller, and $A_{\text{surf}}$ represents the UAV body surface area.

The cumulative UAV computational energy consumption $E_{\text{cmp}}(i,t)$ can be represented as:
\begin{equation}
E_{\text{cmp}}(i, t)=\int_{0}^{t} P_{\text{cmp}} \mathrm{d} t,
\end{equation}
where $P_{\text{cmp}}$ is a constant representing UAV computational power.
\begin{equation}
    P_{\text{cmp}} = P_{\text{static}} + P_{\text{dynamic}},
\end{equation}
\begin{equation}
P_{\text{dynamic}} \approx C^* \cdot V^2 \cdot f^* \cdot \alpha^*,
\end{equation}
where $P_{\text{static}}$ represents the static power of the computing module, $P_{\text{dynamic}}$ represents the dynamic power of the computing module, $C^*$ represents the load capacitance, $V$ represents voltage, $f^*$ represents clock frequency, and $\alpha^*$ represents the activity factor.

The cumulative UAV communication energy consumption $E_{\text{comm}}(i,t)$ can be represented as:
\begin{equation}
E_{\text{comm}}(i, t)=\int_{0}^{t} P_{\text{comm}} (i, t)\mathrm{d} t,
\end{equation}
\begin{equation}
P_{\text {comm }}(i, t)=P_{ut}+P_{ur},
\end{equation}
where $P_{\text{comm}}(i, t)$ represents the communication power of $\text{UAV}_i$ at time $t$. $P_{ut}$ is the transmission power from the UAV to the BSs for localization and data transmission, and the second part $P_{ur}$ is \textcolor{black}{transmission power and received signal power at the UAV receiver}.

As for the monitoring $\text{UAV}_{\text{m},i}$, the energy consumption $E_\text{m}(i,t)$ can be represented as:

\begin{equation}
\begin{split}
     E_\text{m}(i,t)= E_{\text{fly}}(i,t) +& E_{\text{comm}}(i,t) + E_{\text{cmp}}(i,t)+E_{\text{cam}}(i,t), 
\end{split}
\end{equation}
where $E_{\text{cam}}(i,t)$ represents the camera energy consumption, which can be represented as:
\begin{equation}
E_{\text{cam}}(i, t)=\int_{0}^{t} P_{\text{cam}} \mathrm{d} t,
\end{equation}
where $P_{\text{cam}}$ represents the camera power.

As for the data collection $\text{UAV}_{\text{d},i}$, the energy consumption $E_\text{d}(i,t)$ can be represented as:

\begin{equation}
\begin{split}
     E_\text{d}(i,t)= E_{\text{fly}}(i,t) +& E_{\text{comm}}(i,t) + E_{\text{cmp}}(i,t)+E_{\text{ec}}(i,t), 
\end{split}
\end{equation}
where $E_{\text{et}}(i,t)$ represents the extra communication energy consumption for data collection, which is represented as:
\begin{equation}
E_{\text{ec}}(i, t)=\int_{0}^{t} P_{\text{ec}} \mathrm{d} t,
\end{equation}
where $P_{\text{ec}}$ represents the extra communication power.

\subsection{Data Collection Model}
As a metric reflecting data freshness, the AoI for each $\text{WS}_j$ in the WSN at time $t$ is expressed as $\text{AoI}_{j}(t)$, defined by

\begin{equation}
\begin{split}
    \text{AoI}_{j}(t) =
\begin{cases}
0, \text{    }  \text{if  } T_S(j,t)=1,\\
\text{AoI}_{j}(t-t_{\text{step}}), \text{    }\text{if  } T_S(j,t)=0 \text{ and } t \nmid t_{\text{cycle}} \\
\min\{\text{AoI}_{j}(t-t_{\text{step}})+1,\text{AoI}_{\max}\},  \text{otherwise,}
\end{cases}
\end{split}
\label{eq:AoI}
\end{equation}
where $T_S(j,t)$ represents the data transmission success status of $\text{WS}_j$ at time $t$, $t_{\text{cycle}}$ represents the data collection cycle of $\text{WS}_j$, and $t_{\text{step}}$ represents the UAV task time step.

During each update interval $T_{\text{UA}}(i_{\text{type}})$,where $i_{\text{type}}$ denotes the WS type index among $I_{\text{type}}$ categories, the AoI value increases incrementally. Although WSs of various types differ in their updating intervals $T_{\text{UA}}(i_{\text{type}})$, they share a common AoI upper limit $\text{AoI}_{\max}$.
At every time slot, each $\text{WS}_j$ initiates a connection request broadcast toward nearby UAVs. Upon establishing a connection, $\text{WS}_j$ transmits its data packets to the connected $\text{UAV}_i$. If the transmission succeeds, $\text{UAV}_i$ replies with an acknowledgment, resetting $\text{AoI}_j(t)$ before the link is released. Otherwise, the transmission attempt times out, and $\text{WS}_j$ will retry in the subsequent cycle. Once the data are successfully received, $\text{UAV}_{\text{d},i}$ forwards the aggregated packets to $\text{UAV}_{\text{c}}$ through its established communication link.
The packet loss rate (PLR) of $\text{UAV}_{\text{d},i}$ during transmission can be modeled as:
\begin{equation}
\text{PLR}(U_{\text{d}}(i,t)) = 1 - (1 - \text{BER}(U_{\text{d}}(i,t)))^{L},
\end{equation}
where $L$ is the packet length, and $\text{BER}(U_{\text{d}}(i,t))$ denotes the bit error rate at location $U_i(t)$. As binary phase shift keying (BPSK) modulation is applied, the $\text{BER}(U_{\text{d}}(i,t))$ is given by
\begin{equation}
\text{BER}\left(U_{\text{d}}(i,t)\right)=Q\left(\sqrt{2 \cdot \operatorname{SINR}\left(U_{\text{d}}(i,t)\right)}\right),
\end{equation}
where $\operatorname{SINR}(U_{\text{d}}(i,t))$ is the signal-to-interference-plus-noise ratio (SINR) at position $U_{\text{d}}(i,t)$, and $Q(x)$ represents the $Q$-function in communication theory, defined as
\begin{equation}
\begin{array}{l}
\operatorname{SINR}\left(U_{\text{d}}(i,t))\right)= 
\frac{\sum_{k, \text{WS}_k \in \mathrm{WS}}^{n} P_{\text{WT}}(t)+G_{\text{WS}}(t)-PL(i,j,t)}{\sum_{k, \text{WS}_k \notin \mathrm{WS}}^{n} P_{\text{WT}}(t)+G_{\text{WS}}(t) -PL(i,j,t)+P_{\text{T}}},
\end{array}
    \label{20}
\end{equation}
\begin{equation}
\begin{array}{l}
Q(x)=\frac{1}{\sqrt{2\pi}}\int_{x}^{\infty} e^{-\frac{t^{2}}{2}},dt \approx \frac{1}{2}e^{-\frac{x^{2}}{2}},
\end{array}
\label{21}
\end{equation}
where $P_{\text{WT}}(t)$ denotes the transmission power of WSs, $G_{\text{WS}}(t)$ is the antenna gain, $PL(i,j,t)$ represents the propagation path loss between $\text{UAV}i$ and $\text{WS}j$, and $P_{\text{T}}$ is the thermal noise power defined by
\begin{equation}
P_{\text{T}} = k_B \times T_K \times Bw,
\end{equation}
where $k_B$ is Boltzmann’s constant, $T_K$ is the absolute temperature in Kelvin, and $Bw$ denotes the communication bandwidth.
The total path loss $PL(i,j,t)$ between $\text{UAV}_{\text{d},i}$ and $\text{WS}j$ can be expressed as
\begin{equation}
\begin{split}
    PL(i,j,t)= PL_{\text{FS}}(Dis(i&,j,t),f_c) + PL_{\text{veg}}(Dis(i,j,t),f_c),
\end{split}
\end{equation}
\begin{equation}
\begin{split}
    PL_{\text{FS}}(Dis(i,j,t),f_c) = 20\log_{10}(Dis(i,j,t)) +& 20\log_{10}(f_c) + 32.44,
\end{split}  
\end{equation}
\begin{equation}
    PL_{\text{veg}}(Dis(i,j,t),f_c) = a_1 \, f_c^{a_2} \, Dis(i,j,t)^{a_3}, 
\end{equation}
where $Dis(i,j,t)$ indicates the distance between $\text{UAV}_{\text{d},i}$ and $\text{WS}_j$ at time $t$, $f_c$ is the carrier frequency, $PL_{\text{FS}}(\cdot)$ is the free-space path loss, and $PL_{\text{veg}}(\cdot)$ is the vegetation-induced path loss~\cite{iturP833}. Parameters $a_1$, $a_2$, and $a_3$ denote the scaling factor, frequency exponent, and distance exponent, respectively.
\subsection{Monitoring Model}
To facilitate statistics, the farmland is divided into $n_{px} \times n_{py}$ small plots, which can be represented as $\mathcal{P} = {\mathcal{P}}_{1,1},\dots,\mathcal{P}_{n_{\text{px},{\text{py}}}}$, as mentioned in Section~\ref{system_overview}. 
The monitoring UAVs are responsible for patrolling these plots and capturing image data for each grid area. To facilitate the computation and assessment of UAV image quality,
each plot is divided into $n_{\text{gx}} \times n_{\text{gy}}$ grids, which can be represented as $G = G_{0,0}, G_{0,1}\dots, G_{n_{\text{gx}},n_{\text{gy}}}$.
The image quality of each grid is determined based on the UAV’s distance and viewing angle relative to the grid. The overall image quality of the plot is then obtained by aggregating the evaluation of all grids. Assume that the distance between gird $G_{x',y'}$ and the closest UAV can be denoted as $d_{G}(x',y')$. We define a distance threshold $d_{\text{th}}$. If the actual average distance $d_{G}(x',y')$ exceeds $d_{\text{th}}$ (i.e., $d_{G}(x',y') > d_{\text{th}}$), the image quality is considered poor; otherwise, it is deemed acceptable.
The image quality of grid $G_{x',y'}$ can be denoted as $\mathcal{Q}(x',y',t)$, which can be represented as

\begin{equation}
    \mathcal{Q}(x',y',t) = \begin{cases}
0, & \text{if  } d_{G}(x',y')<d_{\text{th}},\\
1, &  \text{Otherwise}.
\end{cases}
\end{equation}

In this context, the image quality $\mathcal{Q}_{\text{plot}}(x,y,t)$ of the whole monitoring plot $\mathcal{P}_{x,y}$ at time $t$ can be represented as

\begin{equation}
\label{14}
    \mathcal{Q}_{\text{plot}}(x,y,t)= \frac{\sum_{x'=0}^{n_{\text{gx}}}\sum_{y'=0}^{n_{\text{gy}}} \mathcal{Q}(x',y',t)}{n_{\text{gx}} \cdot n_{\text{gy}}},
\end{equation}

% We define a sensing quality threshold $\mathcal{Q}_{\text{th}}$. If the quality metric $\mathcal{Q}_{\text{plot}}(x,y,t)$ does not exceed this threshold, the monitoring task in $\mathcal{P}_{x,y}$ is considered unsuccessful. 
We define a lowest acceptable quality threshold $\mathcal{Q}_{\text{th}}$ for $\mathcal{Q}_{\text{plot}}(x,y,t)$ to be deemed successful.
The Monitoring success metric $V_S(x,y,i,t)$ of $\text{UAV}_{\text{m},i}$ in plot $\mathcal{P}_{x,y}$ at time $t$ can be represented as

\begin{equation}
    V_S(x,y,i,t) =  \begin{cases}
0, & \text{if  } \mathcal{Q}_{\text{plot}(x,y,t)} < \mathcal{Q}_{\text{th}},\\
1, &  \text{Otherwise}.
\end{cases}
\end{equation}

As a metric reflecting visual data freshness, the VDF, $\text{UAV}_{\text{m},i}$ visiting, for each plot $\mathcal{P}_{x,y}$ at time $t$ is expressed as $\text{VDF}_{x,y}(i,t)$, defined by
\begin{equation} 
\begin{split} 
\text{VDF}_{x,y}(i,t) =
\begin{cases} 0, & \text{if } \mathcal{Q}_{\text{plot}(x,y,t)} \geq
 \mathcal{Q}_{\text{th}},\\ 
 \text{VDF}_{x,y}(t-t_{\text{step}}) &\text{if} \text{ }\mathcal{Q}_{\text{plot}(x,y,t)} <
 \mathcal{Q}_{\text{th}} \text{ and } t \nmid t_{\text{v}} , \\
 \min\{\text{VDF}_{x,y}(t-t_{\text{step}})+1,\text{VDF}_{\max}\}, & \text{otherwise,} 
\end{cases} 
\end{split} 
\label{eq:VDF}
\end{equation}
where $t_{\text{v}}$ represents the VDF update cycle.

\section{Design of EIA-SEC}
\label{algorithm}
% This section details the Markov Decision Process formulation and proposed algorithms.
\subsection{Markov Decision Process}
% To solve Pro~\eqref{1-a}, \eqref{2-a} and \eqref{3-a} through MARL, a multi-agent MDP is designed as follows:

\subsubsection{Agent} In the considered scenario, each agent is a UAV that makes its own decision. Assume that the agent represents the communication $\text{UAV}_{\text{c}}$ as $\{ \text{agent}_{\text{c},1}, \dots, \text{agent}_{\text{c},n_{\text{c}}} \}$, monitoring $\text{UAV}_{\text{m}}$ as $\{ \text{agent}_{\text{m},1}, \dots, \text{agent}_{\text{m},n_{\text{m}}} \}$, and $\text{UAV}_{\text{d}}$ as $\{ \text{agent}_{\text{d},1}, \dots, \text{agent}_{\text{d},n_{\text{d}}} \}$.
% \subsubsection{Observation of Each Agent $\bm{o}_{\text{c}-i,k}$, $\bm{o}_{\text{m}-i,k}$, and $\bm{o}_{\text{d}-i,k}$}
\subsubsection{Observation of Each Agent}
In the intelligent farming scenario, $\text{UAV}_{\text{c}}$, $\text{UAV}_{\text{m}}$ and $\text{UAV}_{\text{d}}$ have distinct observation spaces. As for $\text{UAV}_{\text{c}}$, the observation space should provide information on the $\text{UAV}_{\text{c}}$ position  $\bm{U}_{\text{c}}(i,k)$; the distance $d_{\text{C-c}}(i,k)$ and coordinates $\text{UAV}_{\text{C-c}}(i,k)$ of the nearest communication UAV; the geometric center $\text{GC}_{\text{c}}(i,k)$ of the $\text{UAV}_{\text{c}}$, and its distance $d_{\text{GC-c}}(i,k)$ to the geometric center; the geometric center $\text{GC}_{\text{u}}(i,k)$ of the served UAV swarm, and its distance  $d_{\text{GC-u}}(i,k)$ to the geometric center; and the distance $d_{\text{f-s}}(i,k)$ and coordinates $\text{UAV}_{\text{f-s}}(i,k)$ of the farthest served UAV; the nearest $\text{UAV}_{\text{N}}(i,k)$ and the distance $d_U(i,k)$ between them.
\begin{equation}
\begin{split}
\bm{o}_{\text{c}-i,k}=\{\bm{U}_{\text{c}}(i,k),  \text{UAV}_{\text{C-c}}(i,k), d_{\text{C-c}}(i,k), \text{GC}_{\text{c}}(i,k), \\d_{\text{GC-c}}(i,k),  \text{GC}_{\text{u}}(i,k), d_{\text{GC-c}}(i,k),
 \text{UAV}_{\text{f-s}}(i,k), &d_{\text{f-s}}(i,k), \text{UAV}_\text{N}(i,k), d_U(i,k)\}, 
\end{split}
\end{equation}

As for $\text{UAV}_{\text{m}}$, the observation space should provide information on UAV position $\bm{U}_{\text{m}}(i,k)$; the VDF information $\{ \text{VDF}_{x-1,y-1}(i,k), \dots,\text{VDF}_{x+1,y+1}(i,k) \}$ around the plot $\mathcal{P}_{x,y}$, where $\text{UAV}_{\text{m},i}$ is flying; the direction $D_{\text{VDF}}(i,k)$ of the quadrant with the highest VDF concentration; the nearest $\text{UAV}_{\text{N}}(i,k)$ and the distance $d_U(i,k)$ between them.
\begin{equation}
\begin{split}
\bm{o}_{i,k}=\{\bm{U}_{\text{m}}(i,k), \text{VDF}_{x-1,y-1}(i,k), \dots,\text{VDF}_{x+1,y+1}(i,k) ,D_{\text{VDF}}(i,k),\text{UAV}_\text{N}(i,k), d_U(i,k)\},
\end{split}
\end{equation}

As for $\text{UAV}_{\text{c}}$, the observation space should provide information on UAV position $\bm{U}_{\text{d}}(i,k)$; the AoI information of nearby WSs $\{ \text{AoI}(i,j_1,k),..., \text{AoI}(i,j_n,k)\}$; the direction $D_{\text{AoI}}(i,k)$ of
the quadrant with the highest AoI concentration; 
the nearest $\text{UAV}_{\text{N}}(i,k)$ \textcolor{black}{on the same flight plane and the distance $d_U(i,k)$ between them.}

\begin{equation}
\begin{split}
\bm{o}_{i,k}=\{\bm{U}_{\text{d}}(i,k), \text{AoI}(i,j_1,k),...,\text{AoI}(i,j_n,k), D_{\text{AoI}}(i,k),
 \textcolor{black}{\text{UAV}_\text{N}(i,k), d_U(i,k)\}},
\end{split}
\end{equation}

% \subsubsection{Action of Each Agent $\bm{a}_{\text{c}-i,k}$, $\bm{a}_{\text{m}-i,k}$, and $\bm{a}_{\text{d}-i,k}$.}

\subsubsection{Action of Each Agent.}
Action space of $\text{UAV}_{\text{c}}$, $\text{UAV}_{\text{m}}$, and $\text{UAV}_{\text{d}}$ are the same, all consist of the flying speed vector
\begin{equation}
\begin{split}
 \bm{a}_{\text{c}-i,k}= \{ v_{\text{c}-x}(i,k), v_{\text{c}-y}(i,k) \}, 
\end{split}
\end{equation}
\begin{equation}
\begin{split}
 \bm{a}_{\text{m}-i,k}= \{ v_{\text{m}-x}(i,k), v_{\text{m}-y}(i,k) \}, 
\end{split}
\end{equation}
\begin{equation}
\begin{split}
 \bm{a}_{\text{d}-i,k}= \{ v_{\text{d}-x}(i,k), v_{\text{d}-y}(i,k) \}, 
\end{split}
\end{equation}
where the speed in $x$ and $y$ direction are within $V_{x\text{-}\max}$ and $V_{y\text{-}\max}$, respectively.

\subsubsection{Reward of Each Agent $r_{\text{c}-i,k}$, $r_{\text{m}-i,k}$, and $r_{\text{d}-i,k}$}
The reward is the feedback provided by the environment to agents based on their observation and chosen action.

As for communication UAVs, we consider to maximize the quality of the communication service $\text{UAV}_{\text{c}}$ providing for $\text{UAV}_{\text{m}}$ and $\text{UAV}_{\text{d}}$ while minimizing waste of resources, boundary violation and collision risk penalties. 

To ensure that the communication $\text{UAV}_{\text{c},i}$ can provide reliable communication services for other task UAVs, we design a communication positioning function $\text{DPF}_{C-i}(t)$ based on the relative positions of the UAVs, which is represented as
\begin{equation}
\begin{split}
    \text{DPF}_{C-i}(t) = \int_{0}^{t}  D_{\text{c}}(i,t) \mathrm{d}t = \sum_{k=0}^
    {k_{\text{t}}} (D_{\text{c}}(i,k+1)& - D_{\text{c}}(i,k)) \cdot t_{\text{step}}.
\end{split}
\end{equation}

The total energy consumption $E_{C-i}(t)$ of $i$-th UAV can be denoted as:
\begin{equation}
    E_{C-i}(t) = E(i,t)  \approx \sum_{k=0}^{k_{\text{t}}} (E(i,k+1)-E(i,k)) \cdot t_{\text{step}}.
\end{equation}

To avoid collisions with each other, a safety risk penalty
$D_{U-i}(t)$ is set to describe the collision risk between $\text{UAV}_{\text{c},i}$ and other UAVs. It can be represented as:

\begin{equation}
    D_{U-i}(t) =\int_{0}^{t} SR(i,t)\mathrm{d}t = \sum_{k=0}^{k_{\text{t}}} SR(i,k) \cdot t_{\text{step}}
\end{equation}

\begin{equation}
    SR(i,t) = \left\{\begin{array}{ll}
d_U(i,t) - d_U(i,t-t_{\text{step}}), & d_U(i,t) < d_{\text{safety}}, \\
0, & \text { otherwise,}
\end{array}\right.
\end{equation}

\begin{equation}
    d_{\text{safety}} = 2 \cdot \sqrt{(V_{x-\max}^2 +V_{y-\max}^2 )} ,
\end{equation}
where $d_U(i,t)$ represents the distance between $i$-th UAV and the closest $\text{UAV}_\text{N}(i,t)$ at time $t$, $SR(i,t)$ represents the safety risks of UAV collisions, and $d_{\text{safety}}$ is the safety threshold.

Besides, to avoid $\text{UAV}_i$ flying out task space, a \textcolor{black}{boundary violation} penalty $\mathcal{P}_{O-i}(t)$ is set as:
\begin{equation}
\mathcal{P}_{O-i}(t) = \int_{0}^{t} ACT(i,t)\mathrm{d}t = \sum_{k=0}^{k_{\text{t}}} ACT(i,k) \cdot t_{\text{step}},
\end{equation}
\begin{equation}
ACT(i, t)=\left\{\begin{array}{ll}
1, & \text { out of range}, \\
0, & \text { otherwise, }
\end{array}\right.
\end{equation}
where $ACT(i,t)$ represents the area compliance tracking of $i$-th UAV at time $t$.

The communication UAV reward can be represented as
\begin{equation}
\begin{split}
    r_{\text{c}-i,k}  = [\alpha_1 (D_\text{c}(i,k)-D_\text{c}(i,k-1))
    -\alpha_2 (E(i,k)-E(i,k-1))
    -\alpha_3 SR(i,k) - \alpha_4 ACT(i,k)] \cdot t_{\text{step}}. 
\end{split}  
\end{equation}

As for monitoring UAVs, we consider maximizing the quality of the monitoring task of $\text{UAV}_{\text{m}}$ while minimizing waste of resources, boundary violations, collision risks, and capture failure penalties. 

To motivate the UAVs to capture efficiency, we describe visual monitoring utility value $\text{VMUF}(i,t)$ as
\begin{equation}
     \text{VMUF}(i,t)=\int_{0}^{t} \text{VDF}_{x,y}(i,t) \cdot  V_S(x,y,i,t)\,\mathrm{d}t, 
\end{equation}

To evaluate the quality of monitoring of $\text{UAV}_{\text{m},i}$, we define a VDF penalty $\mathcal{P}_{M}(i,t)$ as
\begin{equation}
\mathcal{P}_{M}(i,t) =
\begin{cases}
1, & \text{if }  
\text{VMUF}(i,t)=0,\\
0, & \text{otherwise}.
\end{cases}
\end{equation}

To ensure the trend of efficient monitoring, we define a VDF motivation $\mathcal{M}_{V}(i,t)$ as
\begin{equation}
    \mathcal{M}_{V}(i,t) = \begin{cases}
        1, & \text{if } \arccos\!\left( \frac{\bm{v}_{\text{m}}(i,t)\cdot D_{\text{VDF}}(i,t)}{\|\bm{v}_{\text{m}}(i,t)\|\|D_{\text{VDF}}(i,t)\|} \right) < 90^\circ \\
        0, &\text{otherwise}

    \end{cases},
\end{equation}

We design the $r_{\text{m}-i,k}$ as
\begin{equation}
\begin{split}
    r_{\text{m}-i,k}  = [- \alpha_2( E(i,k) -E(i,k-1) )-\alpha_3 SR(i,k) - \alpha_4 ACT(i,k)\\+\alpha_5\text{VMUF}(i,k) 
    - \alpha_6 \mathcal{P}_{M}(i,k) 
    +\alpha_7\mathcal{M}_{V}(i,k)] \cdot t_{\text{step}}.  
\end{split}  
\end{equation}

As for data collection UAVs, we consider to maximize utility value of collected data of $\text{UAV}_{\text{d}}$ while minimizing waste of resources, boundary violation, collision risk, and waste of resources penalties.

To motivate the UAVs to collect rich and up-to-date data, we denote the data collection utility function $ \text{DCUF}(i,t)$ as
\begin{equation}
     \text{DCUF}(i,t)=\int_{0}^{t} \text{AoI}(i,j,t) \cdot T_S(i,j,t)\,\mathrm{d}t, 
\end{equation}
where $\text{AoI}(i,j,t)$ denotes the AoI of $\text{WS}_j$ that $\text{UAV}_{\text{d},i}$ collects at time $t$ and $T_S(i,j,t)$ represents the data transmission success status function.

To evaluate the quality of data collection of $\text{UAV}_{\text{d},i}$, we define an AoI penalty $\mathcal{P}_{C}(i,t)$ as
\begin{equation}
\mathcal{P}_{C}(i,t) =
\begin{cases}
1, & \text{if }  % \text{AoI}(i,j,t) = 0
\text{DCUF}(i,t)=0,\\
0, & \text{otherwise},
\end{cases}
\end{equation}

We define an AoI motivation $\mathcal{M}_{d}(i,t)$ as
\begin{equation}
    \mathcal{M}_{d}(i,t) = \begin{cases}
        1, & \text{if } \arccos\!\left( \frac{\bm{v}_{\text{d}}(i,t)\cdot D_{\text{AoI}}(i,t)}{\|\bm{v}_{\text{d}}(i,t)\|\|D_{\text{AoI}}(i,t)\|} \right) < 90^\circ \\
        0, &\text{otherwise}
    \end{cases},
\end{equation}

We design the  $r_{\text{d}-i,k}$ as
\begin{equation}
\begin{split}
    r_{\text{d}-i,k}  = [- \alpha_2( E(i,k) -E(i,k-1) )-\alpha_3 SR(i,k) - \alpha_4 ACT(i,k)\\
    +\alpha_8\text{DCUF}(i,k) 
    - \alpha_9 \mathcal{P}_{C}(i,k) +\alpha_{10}\mathcal{M}_{D}(i,k)] \cdot t_{\text{step}}.  
\end{split}  
\end{equation}

\subsection{Elite Imitation Actor}
\label{eia}
To expedite the training process and enable agents to leverage the behavioral patterns of high-performing individuals, we introduce an elite imitation mechanism. This mechanism is formulated to mitigate prolonged low-reward exploration by guiding agents toward more effective behavioral trajectories observed in the selected elite.

We encourage agents belonging to the same category to learn from one another. Let $\text{agent}_{\text{v}}$ denote a generic agent category that shares comparable behavioral patterns and underlying task structures, which enables more effective intra-category knowledge transfer, where $\text{v} =  \{\text{c},\text{m},\text{d}\}$.
We introduce a mimicry cycle factor, denoted by $\delta_{\text{v}}$. During certain episodes determined by this factor, each $\text{agent}_{\text{v-}i}$ temporarily suspends the direct update of its policy network $\pi_{\text{v-}i}(\theta_
{\text{v-}i})$. The agent generates an action sequence 
$\mathcal{A}_{\text{v-}i} = \{ a_{\text{v-}i,0}, a_{\text{v-}i,1}, \dots, a_{\text{v-}i,n_{\text{v}}} \}$ 
and interacts with the environment to obtain a corresponding reward vector 
$\mathcal{R}_{\text{v-},i} = \{ r_{\text{v-}i,1}, r_{\text{v-}i,2}, \dots, r_{\text{v-}i,n_{\text{v}}} \}$, where $n_{\text{v}}$ represents the number of $\text{agent}_{\text{v}}$.

The elite policy is then identified based on the mean $\mu_{\text{v-}i}$ and variance $\sigma_{\text{v-}i}^2$ of the rewards in $\mathcal{R}_{\text{v-}i}$. 
To quantify the performance of $\text{agent}_{\text{v-}i}$ , we define an evaluation metric as:
\begin{equation}
\omega_{\text{v-}i} = \beta_1 \mu_{\text{v-}i} + \beta_2\sigma_{\text{v-}i}^2.  
\label{er}
\end{equation}

Specifically, we select the $\text{agent}_{\text{v-}i'}$ with the best-performing reward profile as the elite individual. 
Subsequently, the remaining agents update their policy networks through a soft imitation process, 
where parameters are softly copied from the elite policy using a soft update coefficient $\vartheta$. 
The update rule is formulated as:

\begin{equation}
    \theta_{\text{v-}i} = (1-\vartheta)\theta_{\text{v-}i} +\vartheta\theta_{\text{v-}i'} ,
    \label{soft}
\end{equation}
where $\theta_{\text{v-}i'}$ and $\theta_{\text{v-}i}$ denote the parameters of the elite and agents’ policy networks, respectively.
Furthermore, to preserve a certain level of exploration while leveraging the elite imitation strategy, we progressively adjust the influence of the imitation process over time. 
Specifically, the soft update coefficient $\vartheta$ is gradually decreased, whereas the mimicry cycle factor $\delta$ is progressively increased. We initialize these parameters as $\vartheta^{*}$ and $\delta^{*}$, respectively. 
The detailed procedure of the proposed algorithm is presented in \textbf{Algorithm~\ref{Algorithm1}}.
\begin{algorithm}[H]
    \caption{Elite Imitation Actor Mechanism}
    \label{Algorithm1}
    \begin{algorithmic}[1]
        \STATE Initialize policy network $\pi_{\text{v-}i}(\theta_{\text{v-}i})$ for each $\text{agent}_{\text{v-}i}$.
        \STATE Initialize $\vartheta \leftarrow \vartheta^{*}$ and $\delta \leftarrow  \delta^{*}$.
        \FOR{episode $\lambda = 1, 2, \dots, \lambda_{\text{max}}$}
        \IF{$\lambda \mid \delta $}
        \FOR{ $\text{agent}_{\text{v-}i} \text{ } i = 1, 2, \dots, n_{\text{v}} $}
        \STATE Generate $\mathcal{A}_{\text{v-}i} $ from $\pi_{\text{v-}i} (\theta_{\text{v-}i} )$. 
        \STATE Obtain Reward $\mathcal{R}_i$ from the interaction between $\text{agent}_{\text{v-}i}$ and the environment.
        \ENDFOR
         \STATE Compute the maximum reward evaluation parameter $\omega_{\text{v-}i} $ as Equation~\eqref{er}.
        \STATE Update $\pi_{\text{v-}i} (\theta_{\text{v-}i} )$ as Equation~\eqref{soft}.
        \STATE Update parameters $\vartheta \leftarrow  \frac{1}{2}\vartheta$ and $\delta \leftarrow  2\delta$.
        \ENDIF
        \ENDFOR 
    \end{algorithmic}
\end{algorithm}

\subsection{Shared Ensemble Critic}
\label{sec}
To enable the critics to provide more impartial evaluations of action-state values, we introduce the SEC mechanism. Similarly, we define $\text{agent}_{\text{v}}$ as a generic category of agents that share behavioral patterns and underlying task structures, thereby facilitating more effective intra-category knowledge transfer, where $\text{v} =  \{\text{c},\text{m},\text{d}\}$.
The SEC mechanism ensures that each $\text{agent}_{\text{v-}i}$ is equipped with its own independent critic $\mathcal{Q}_{\text{v-}i}(\phi_{\text{v-}i})$ and target critic, $\mathcal{Q}'_{\text{v-}i}(\phi'_{\text{v-}i})$, while a shared ensemble of critics $\mathcal{Q}_{\text{s,v}}(\phi_{\text{s}}) = \{ \mathcal{Q}_{\text{s,v-}1}(\phi_{\text{s-1}}),\dots,\mathcal{Q}_{\text{s,v-}k_{\text{s}}}(\phi_{\text{s,v-}k_{\text{s}}})\}$ is additionally constructed for the entire $\text{agent}_{\text{v}}$ category to enhance stable value estimation across agents.

Assuming that during the update, $R_{B,\text{v-}i}$ represents a batch size of reward from $\text{agent}_{\text{v-}i}$, the target value can be represented as:
\begin{equation}
    \mathcal{Y} = R_{B,\text{v-}i} + \gamma \cdot  \mathcal{Q}'_{\text{v-}i}(\phi'_{\text{v-}i})
\end{equation}
where $\gamma$ represents the discount rate. 

Then we define the predicted Q-value $\mathcal{Q}_{P,\text{v-}i}$ as:
\begin{equation}
    \mathcal{Q}_{P,\text{v-}i}  = \epsilon \cdot \mathcal{Q}_{\text{v-}i}(\phi_{\text{v-}i}) +  (1-\epsilon) \cdot (\sum_{k=0}^{k_{\text{s}}}\frac{\mathcal{Q}_{\text{s,v-}k}(\phi_{\text{s-}k})}{k_{\text{s}}})
    \label{value}
\end{equation}
where $\epsilon$ represents the rate parameter.
We minimize the mean squared error (MSE) between the predicted Q-values and target Q-values. We update the critic and SEC networks by minimizing the loss function $\mathcal{J}_{\mathcal{Q}}(\phi_{\text{v-}i})$:
    \begin{equation}
    \mathcal{J}_{\mathcal{Q}_{\text{v-}i}}(\phi_{\text{v-}i}) = \mathcal{J}_{\mathcal{Q}_{\text{s,v-}k}}(\phi_i) = \mathbb{E}[(\mathcal{Q}_{P,\text{v-}i}-\mathcal{Y})^2],
    \label{up1}
    \end{equation}
where $\mathbb{E}(\cdot)$ represents the the mathematical expectation.

Moreover, to ensure sufficient diversity among the ensemble members, we proportionally replicate the parameters of the Q-networks within the SEC as
\begin{equation}
    \phi_{\text{s,v-}k} = \frac{\tau k}{k_{\text{s}}}  \phi_{\text{v-}i} + (1-\frac{\tau k}{k_{\text{s}}} )\phi_{\text{s,v-}k},
    \label{up2}
\end{equation}
where $\tau$ represents the parameter rate.
Also, we will softly update the target critic networks toward the current critic networks to ensure training stability. Instead of directly copying the weights, the target networks are updated using a weighted average of the current and previous target weights:
\begin{equation}
    \phi'_{\text{v-}i} = \xi \phi_{\text{v-}i}+(1-\xi)\phi'_{\text{v-}i}
\label{xt}
\end{equation}
where $\xi$ is the soft update parameter.

The detailed procedure of proposed algorithm is presented in \textbf{Algorithm~\ref{Algorithm2}}.
\begin{algorithm}[H]
    \caption{Shared Ensemble Critics Mechanism}
    \label{Algorithm2}
    \begin{algorithmic}[1]
        \STATE Initialize critic network $\mathcal{Q}_{\text{v-}i}(\phi_{\text{v-}i})$ for each $\text{agent}_{\text{v-}i}$.
        \STATE Initialize $\mathcal{Q}_{\text{s,v}}(\phi_{\text{s}}) = \{ \mathcal{Q}_{\text{s,v-}1}(\phi_{\text{s-1}}),\dots,\mathcal{Q}_{\text{s,v-}k_{\text{s}}}(\phi_{\text{s,v-}k_{\text{s}}})\} $.
        \FOR{episode $\lambda = 1, 2, \dots,\lambda_{\max}$}
        \FOR{ $\text{agent}_{\text{v-}i} \text{ } i = 1, 2, \dots, n_{\text{v}} $}
        \STATE Generate action by policy network $\pi_{\text{v-}i}(\theta_{\text{v-}i})$.
        \STATE Evaluate the action-state value by \eqref{value}.
        \STATE Update $\mathcal{Q}_{\text{v-}i}(\phi_{\text{v-}i})$ by \eqref{up1}.
        \STATE Update $\mathcal{Q}_{\text{s,v}}(\phi_{\text{s}})$ by \eqref{up2}.
        \STATE Softly update $\mathcal{Q'}_{\text{v-}i}(\phi'_{\text{v-}i})$ by \eqref{xt}.
        \ENDFOR
        \ENDFOR 
    \end{algorithmic}
\end{algorithm}

\begin{figure}[H]
    \centering
    \includegraphics[width=0.9\textwidth, keepaspectratio]{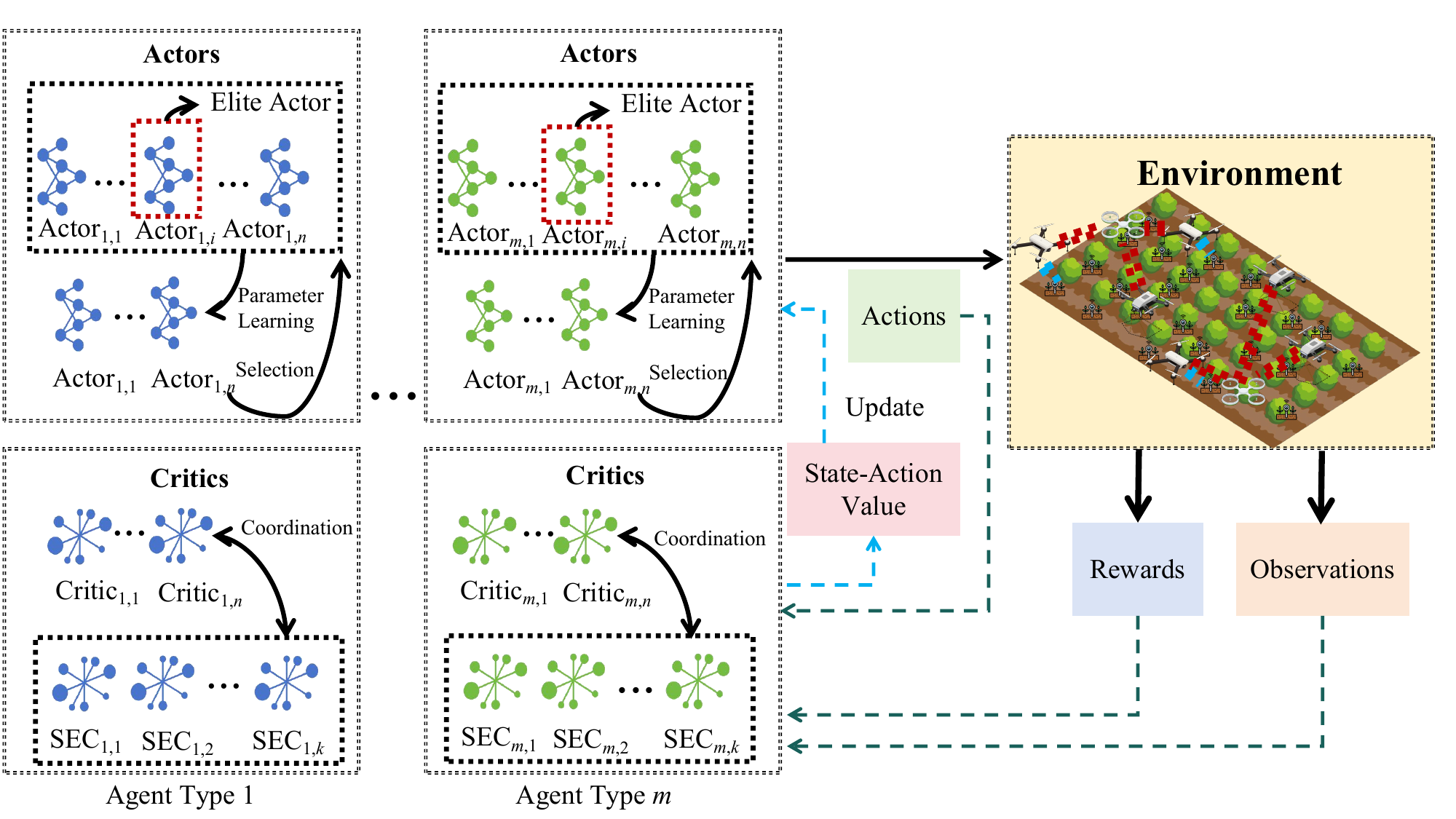}
    \caption{The structure of our EIA-SEC framework.}
\label{fig:eia-sec}
    \setlength{\belowcaptionskip}{-10pt}  % 调整标题下方间距
\end{figure}
Assume that $o_{\text{v-}i}$, $a_{\text{v-}i}$, $r_{\text{v-}i}$, and $o_{\text{v-}i+1}$ denote the observation, action, reward, and next observation of $\text{agent}_{\text{v-}i}$, respectively. Let $S_{B,\text{v-}i}$ and $A_{B,\text{v-}i}$ denote the sampled state and action batches, and $\mathcal{D}$ the replay buffer. We update actor $\mathcal{J}_{\mathcal{\pi}_{\text{v-}i}}(\theta_{\text{v-}i}|A_{B,\text{v-}i},S_{B,\text{v-}i})$ by maximizing the sum of the estimated Q-values from the critic $\mathcal{Q}_{\text{v-}i}(\phi_{\text{v-}i}|A_{B,\text{v-}i},S_{B,\text{v-}i})$. The loss function $\mathcal{J}_{\mathcal{\pi}_{\text{v-}i}}(\theta_{\text{v-}i})$ can be represented as:
    \begin{equation}
    \mathcal{J}_{\mathcal{\pi}_{\text{v-}i}}(\theta_{\text{v-}i})= \mathbb{E}[\mathcal{Q}_{\text{v-}i}(S_{B,\text{v-}i},\pi_{\theta_{\text{v-}i}}(\theta_{\text{v-}i}|A_{B,\text{v-}i},S_{B,\text{v-}i}))].
    \label{up3}
    \end{equation}
    
% The training process of EIA-SEC is shown in \textbf{Algorithm~\ref{algorithm3}}.

\begin{algorithm}
 \caption{The training process of the EIA-SEC.}
 \begin{algorithmic}[1]
\STATE Initialize actors $\mathcal{\pi}_{\text{v-}i}(\theta_{\text{v-}i})$, $\text{v} =  \{\text{c},\text{m},\text{d}\}$, $i=1,...,n_\text{v}$.
\STATE Initialize critics $\mathcal{Q}_{\text{v-}i}(\phi_{\text{v-}i})$, $\text{v} =  \{\text{c},\text{m},\text{d}\}$, $i=1,...,n_\text{v}$.
\STATE Initialize target critics $\mathcal{Q'}_{\text{v-}i}(\phi'_{\text{v-}i})$ as \eqref{xt}, $\text{v} =  \{\text{c},\text{m},\text{d}\}$, $i=1,...,n_\text{v}$.
\STATE Initialize SECs $\mathcal{Q}_{\text{s,v}}(\phi_{\text{s}})$,  $\text{v} =  \{\text{c},\text{m},\text{d}\}$, $i=1,...,n_\text{v}$.
\STATE Reset environment.
\FOR {episode $\lambda = 0,1,\dots,\lambda_{\max}$}
\STATE Update elite imitation strategy as \textbf{Algorithm~\ref{Algorithm1}}.
\FOR {step $\lambda' = 0,1,\dots,\lambda'_{\max}$}
\FOR {$\text{v}=  \{\text{c},\text{m},\text{d}\}$}
\FOR{$\text{agent}_{\text{v-}i} \text{ }i = 1,2,\dots,n_{\text{v}}$}
\STATE Output action $\bm{a}_{\text{v-}i,\lambda'}$ from $\mathcal{\pi}_{\text{v-}i}(\theta_{\text{v-}i})$.
\STATE Interact with environment, store $o_{\text{v-}i}$, $a_{\text{v-}i}$, $r_{\text{v-}i}$, and $o_{\text{v-}i+1}$ to $\mathcal{D}$. 
\STATE Update $\mathcal{Q}_{\text{v-}i}(\phi_{\text{v-}i})$, $\mathcal{Q'}_{\text{v-}i}(\phi'_{\text{v-}i})$, and $\mathcal{Q}_{\text{s,v}}(\phi_{\text{s}})$ as \textbf{Algorithm~\ref{Algorithm2}}.
\STATE Update $\mathcal{\pi}_{\text{v-}i}(\theta_{\text{v-}i})$ as \eqref{up3}.
\ENDFOR
\ENDFOR
\ENDFOR
\ENDFOR

 \end{algorithmic}
 \label{algorithm3}
\end{algorithm}

\subsection{EIA-SEC Framework}

As shown in Fig.~\ref{fig:eia-sec}, we combine the EIA and SEC mechanisms with the actor-critic framework of multi-agent reinforcement learning (MARL) for trajectory optimization problems. In our proposed EIA-SEC framework, each agent cooperates with both the same type and different types of other agents. As mentioned in Section~\ref{eia}, actors periodically select the advantaged individuals for learning, and generate action policies to interact with the environment. The interaction with environments leads to the rewards and states for critics to evaluate the action-state value of the action policy. As introduced in Section~\ref{sec}, agents within the same type, in addition to their own individual critics, are also equipped with SECs, which interact and co-learn with the individual critics. This design helps prevent the overestimation of Q-values and leads to more impartial action-state value evaluations. Moreover, since the SECs are continuously updated through interactions with different agents within the type, they implicitly enable cross-agent critique sharing, thereby accelerating the learning process.

% \begin{figure}[H]
%     \centering
%     \includegraphics[width=0.42\textwidth, keepaspectratio]{fig/EIA-SEC.pdf}
%     \caption{The structure of EIA-SEC MARL.}
% \label{fig:eia-sec}
%     \setlength{\belowcaptionskip}{-10pt}  % 调整标题下方间距
% \end{figure}
Assume that $o_{\text{v-}i}$, $a_{\text{v-}i}$, $r_{\text{v-}i}$, and $o_{\text{v-}i+1}$ denote the observation, action, reward, and next observation of $\text{agent}_{\text{v-}i}$, respectively. Let $S_{B,\text{v-}i}$ and $A_{B,\text{v-}i}$ denote the sampled state and action batches, respectively, and $\mathcal{D}$ denote the replay buffer. We update actor $\mathcal{J}_{\mathcal{\pi}_{\text{v-}i}}(\theta_{\text{v-}i}|A_{B,\text{v-}i},S_{B,\text{v-}i})$ by maximizing the sum of the estimated Q-values from the critic $\mathcal{Q}_{\text{v-}i}(\phi_{\text{v-}i}|A_{B,\text{v-}i},S_{B,\text{v-}i})$. The loss function $\mathcal{J}_{\mathcal{\pi}_{\text{v-}i}}(\theta_{\text{v-}i})$ can be represented as:
    \begin{equation}
    \mathcal{J}_{\mathcal{\pi}_{\text{v-}i}}(\theta_{\text{v-}i})= \mathbb{E}[\mathcal{Q}_{\text{v-}i}(S_{B,\text{v-}i},\pi_{\theta_{\text{v-}i}}(\theta_{\text{v-}i}|A_{B,\text{v-}i},S_{B,\text{v-}i}))].
    \label{up3}
    \end{equation}
    
The training process of EIA-SEC is shown in \textbf{Algorithm~\ref{algorithm3}}.

\section{Evaluation}
\label{simulation_results}
% This section illustrates the experimental setup and results.

\subsection{Experimental Setup} \label{setting}
\begin{table*}[t]
\caption{Environmental Parameters for the Agricultural Application~\cite{zhuozhou}.}
\label{env_agri}
\centering
\begin{tabular}{|c|c|c|c|c|c|c|c|} \hline
\textbf{Symbol} & \textbf{Definition} & \textbf{Value} &  \textbf{Unit} & \textbf{Symbol} & \textbf{Definition} & \textbf{Value} &  \textbf{Unit} \\
\hline
$U_{x-\text{min}}$ & Task Space Left Edge & $0$ & m &$U_{x-\text{max}}$ & Task Space Right Edge & $400$ & m\\
$U_{y-\text{min}}$ & Task Space Back Edge & $0$ & m &$U_{y-\text{max}}$ & Task Space Front Edge & $400$ & m\\
$V_{x-\text{max}}$ & $x$ Dire. Maximum Velocity& 10 & $\text{m/s}$ & $V_{y-\text{max}}$ & $y$ Dire. Maximum Velocity & 10 & $\text{m/s}$ \\
$H_{\text{c}}$ & $\text{UAV}_\text{c}$ Heights  & 22 & m & $H_{\text{m}}$ & $\text{UAV}_\text{m}$ Heights  & 20 & m
\\
,$H_{\text{d}}$ & $\text{UAV}_\text{d}$ Heights  & 18 & m & $n_{\text{WS}}$ & Total WS Number & 400 & m\\
$n_{px}$ & $x$ Dire. Plot Number & 20 & - & $n_{py}$ & $y$ Dire. Plot Number & 20 &-\\
$\text{AoI}_{\text{max}}$ & Maximum AoI & 5 &- &
$\text{VDF}_{\text{max}}$ & Maximum VDF & 5 &- \\
$\epsilon_1$ & Distance Weight Parameter & 0.5& - &
$\epsilon_2$ & Distance Weight Parameter & 0.2& - \\
$\epsilon_3$ & Distance Weight Parameter & 0.3& - &
$m_{\text{UAV}}$ & UAV Mass & 0.2 &kg \\
$g$ & Gravitational Acceleration & 9.8&- & $\rho_{\text{air}}$ & Air Density & 1.225 & $\text{kg/m}^3$\\
$v_{\text{th}}$& Hovering Speed Threshold & 0.1 &$\text{m/s}$ & $C_d$ & Viscosity Coefficient& 0.5 & $-$\\
$n_{\text{prp}}$ &Propeller Number & 4 &- & $R_{\text{prp}}$& Propeller Radius& 0.1 & m\\
$\eta$ &Mechanical Efficiency & 0.8 & - & $A_{\text{surf}}$& UAV Fuselage Area & 0.01 & $\text{m}^2$\\
$P_{\text{static}}$ & Static Power & 4 & W & $V$ & Voltage & 5 & V \\
$f^*$ & Clock Frequency & $200 \cdot  10^6$ & Hz &$\alpha^*$ & Activity Factor & 0.5 & - \\
$C^*$ & Load Capacitance & 6.4 & nF &$T_{\text{update}_{\text{AoI}}}$& AoI Updating Time & 40,50,60& s\\ 
$P_{ur}$ & UAV Received Power & 20 & dBm & $P_{ut}$ & UAV Transmission Power & 20 & dBm \\
$P_{\text{cam}}$ & Camera Power & 2.5 & W & $P_{\text{ec}}$ & Extra Communication Power & 40 & dBm\\
$k_B$ & Boltzmann Constant & $1.38 \cdot 10^{-23}$ & - &
$T_K$ & Temperature (Kelvin) & 298 & K \\
$Bw$ & Bandwidth & 20 &MHz &$f_{c}$ & Signal Frequency & 2.8 & GHz\\ 
$I_{\text{type}}$ & WS Type Number & 3& -& $T_{\text{UA}}$ & AoI Updating Time & 40,50,60& s\\ 
$L$ & Package Length & 20 & Bytes & $t_{\text{step}}$ & Time Cell & 1 & s\\
$t_{\text{v}}$  & VDF Update Cycle & 30 & s & $a_1$ & Scaling Factor & 0.2 &-\\ 
$a_2$ & Frequency Exponent & 0.5 &-& $a_3$ & Distance Exponent & 0.3 &-\\  

\hline
\end{tabular}
\end{table*}

\begin{table*}[t]
\caption{Hyperparameters for the Agricultural Application. }
\label{hyper_agri}
\centering
\begin{tabular}{ |c|c|c|c|c|c|} \hline
\textbf{Symbol} & \textbf{Definition} & \textbf{Value}  & \textbf{Symbol} & \textbf{Definition} & \textbf{Value} \\
\hline
$\alpha_1$ & DPF Parameter &1&$\alpha_2$ & Energy Parameter &0.005\\
$\alpha_3$ & Collision Risky Penalty &5& $\alpha_4$ & Boundary Violation Penalty &10\\
$\alpha_5$ & VMUF Parameter &2& $\alpha_6$ &VDF Penalty &0.5\\
$\alpha_7$ & VDF Motivation &0.2 & $\alpha_8$ & DCUF Parameter &2\\
$\alpha_9$ & AoI Penalty &0.5 & $\alpha_{10}$ & AoI Motivation &0.2\\
$\gamma$ & Discounted Factor& 0.99 & $\xi$ & Soft Update parameter & 0.005\\
$r_a$ & Learning Rate for Actor& $10^{-4}$ &$r_{c}$ & Learning Rate for Critic& $10^{-5}$\\
$\mathcal{D}$ & Replay Buffer Size & $2^{16}$&$\mathcal{B}$ &RL Training Batch Size & 128\\
$\lambda_{\text{max}}$ & Maximum Training Episode & 2000 & $\lambda'_{\text{max}}$ &Maximum Training Step &500\\
$\vartheta^{*}$ & Update Coefficient &0.1 & $\delta^{*}$ & Mimicry Cycle Factor&10\\
$\tau$ & SEC Update Parameter & 0.1 &$\epsilon$ & Rate Parameter& 0.1\\

\hline
\end{tabular}
\end{table*}

\begin{figure*}[t]
    \centering
        \begin{subfigure}{0.28\textwidth}
        \centering
    \includegraphics[width=\textwidth, keepaspectratio]{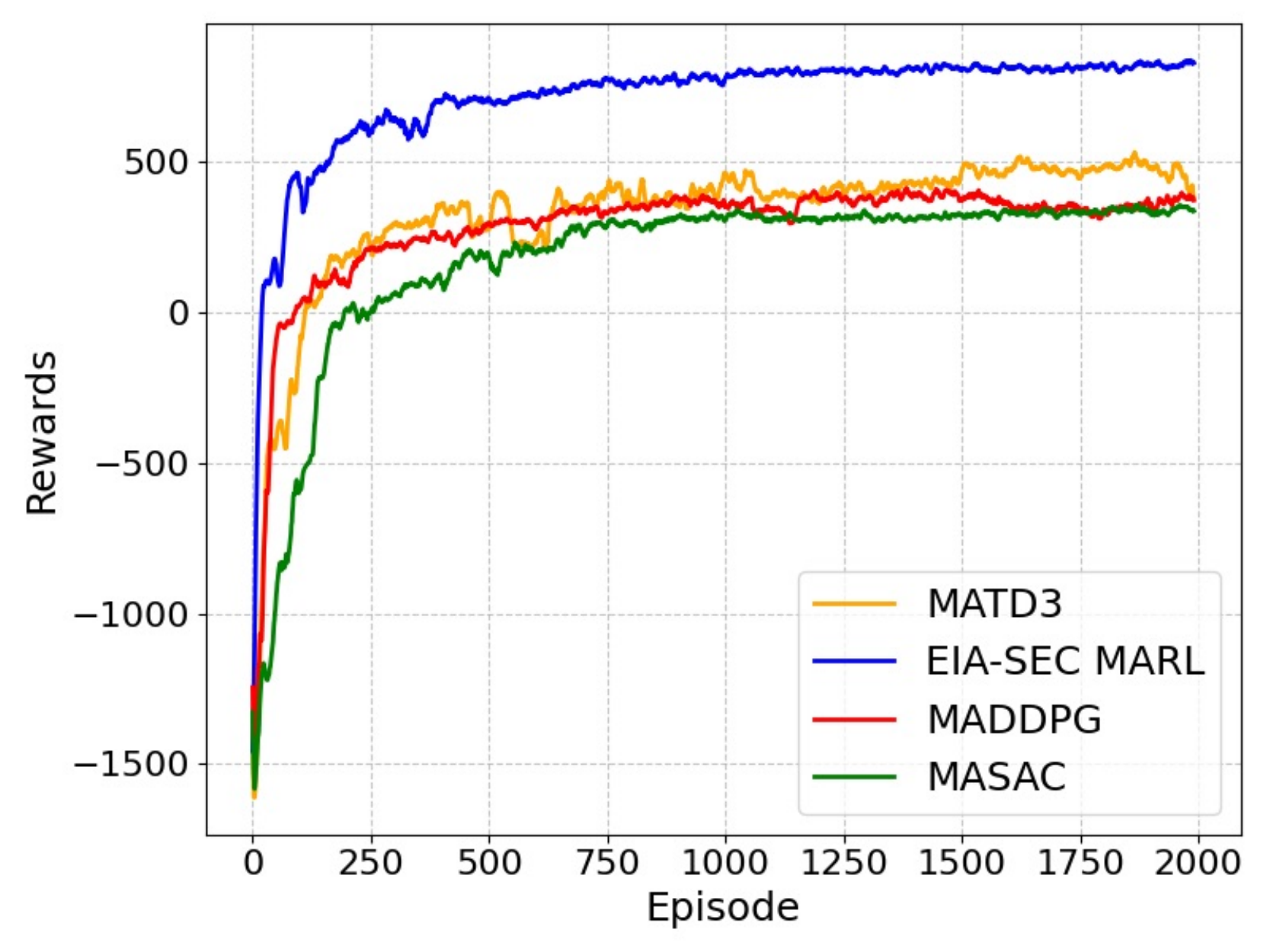}
    \caption{Reward performance comparison with other DRL algorithms.}
\label{fig: reward_performance}
    \setlength{\belowcaptionskip}{-10pt}  
    \end{subfigure}
    \hspace{2em}
    \begin{subfigure}{0.28\textwidth}
        \centering
        \includegraphics[width=\textwidth, keepaspectratio]{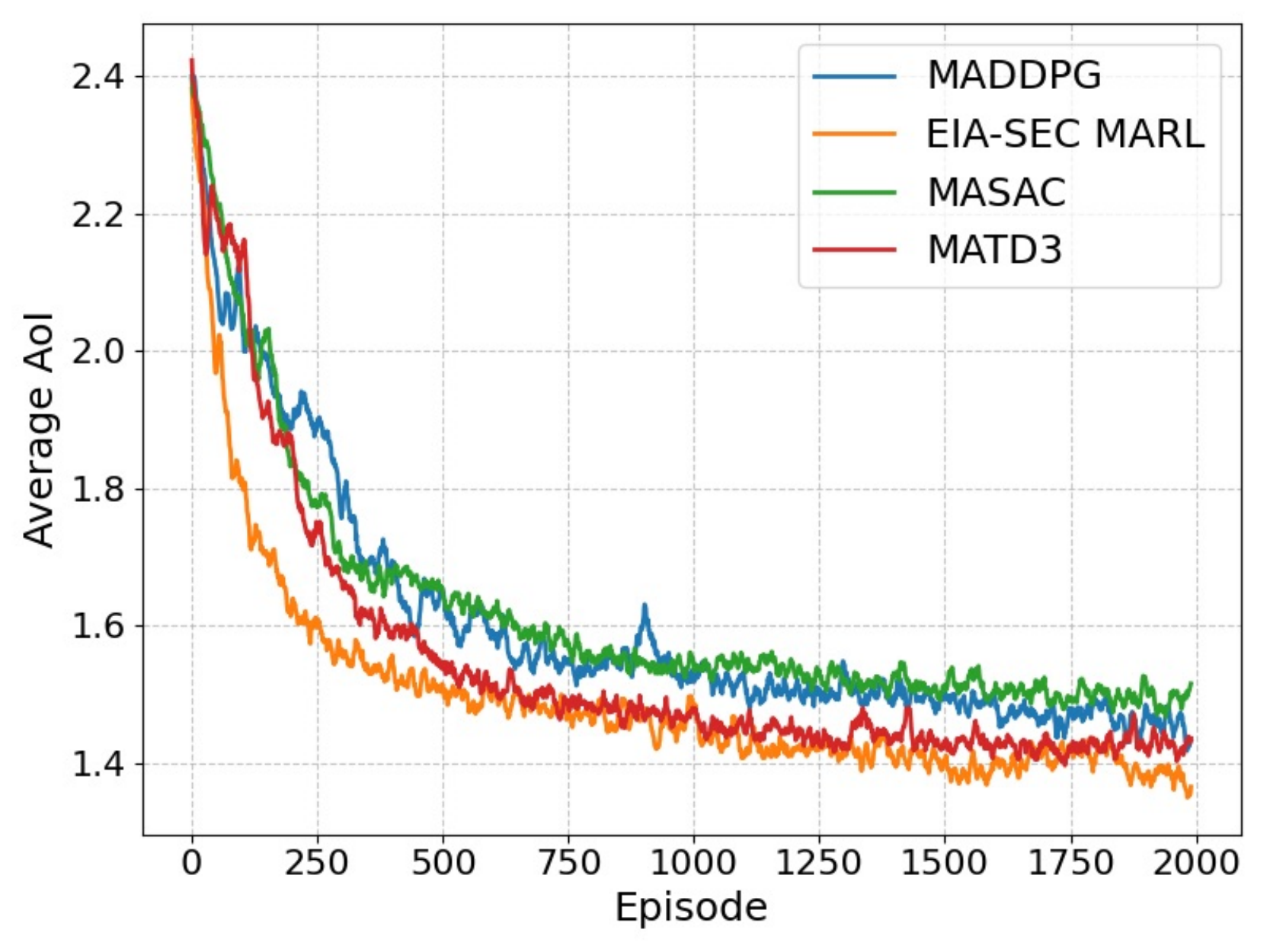}
    \caption{AoI performance comparison with other DRL algorithms.}
    \label{fig:data_aoi}
    \setlength{\belowcaptionskip}{-10pt}
    \end{subfigure}
    \hspace{2em}
    \begin{subfigure}{0.28\textwidth}
        \centering
\includegraphics[width=\textwidth, keepaspectratio]{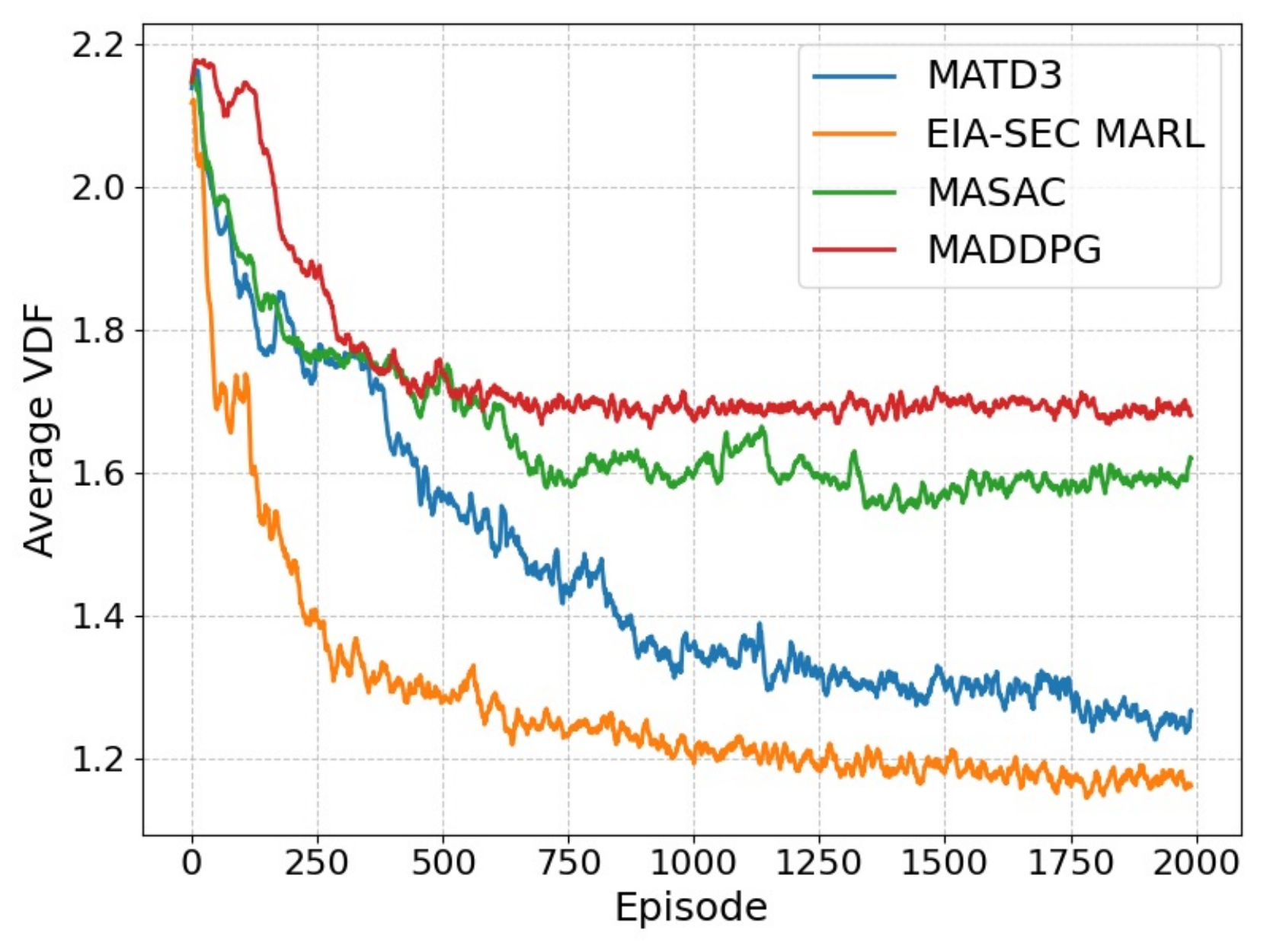}
    \caption{VDF performance comparison with other DRL algorithms.}
    \label{fig:visual_aoi}
    \setlength{\belowcaptionskip}{-10pt}
    \end{subfigure}
        \begin{subfigure}{0.3\textwidth}
        \centering
        \includegraphics[width=\textwidth, keepaspectratio]{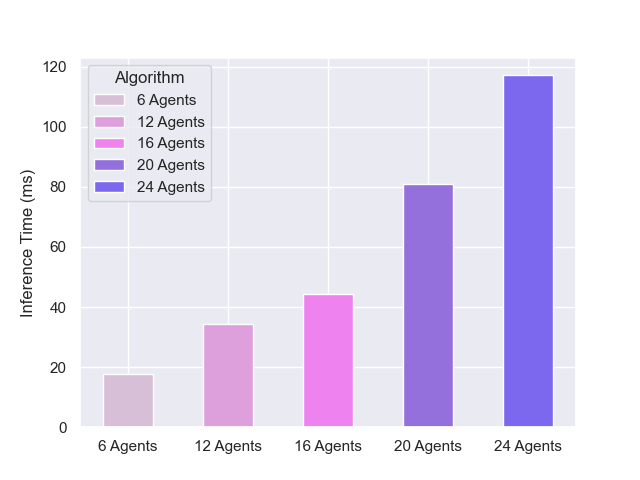}
    \caption{Inference time of EIA-SEC in the scalability study.}
    \label{fig: scalable_in}
    \setlength{\belowcaptionskip}{-10pt}
    \end{subfigure}
           \begin{subfigure}{0.3\textwidth}
        \centering
        \includegraphics[width=\textwidth, keepaspectratio]{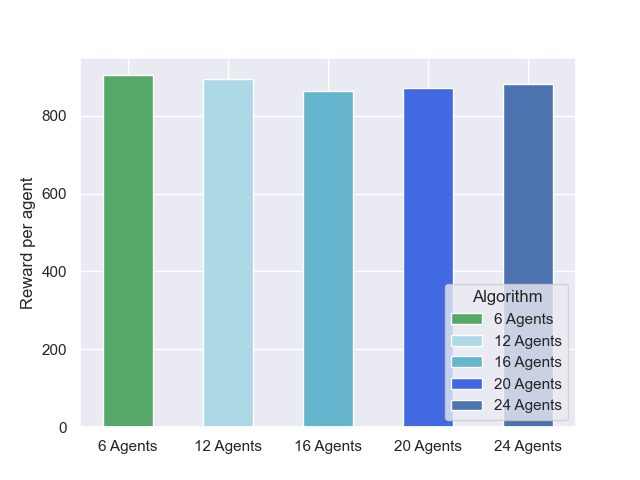}
    \caption{Reward of EIA-SEC in the scalability study.}
    \label{fig: scalable_re}
    \setlength{\belowcaptionskip}{-10pt}
    \end{subfigure}
            \begin{subfigure}{0.3\textwidth}
        \centering
        \includegraphics[width=\textwidth, keepaspectratio]{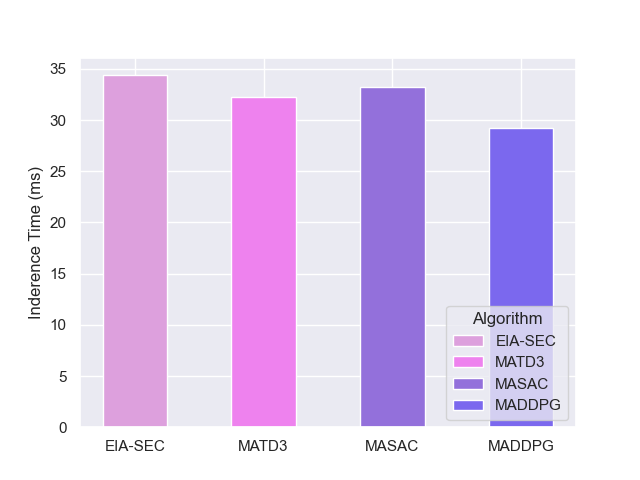}
    \caption{Inference time comparison with other DRL algorithms.}
    \label{fig: scalable_com}
    \setlength{\belowcaptionskip}{-10pt}
    \end{subfigure}
            \begin{subfigure}{0.3\textwidth}
        \centering
        \includegraphics[width=\textwidth, keepaspectratio]{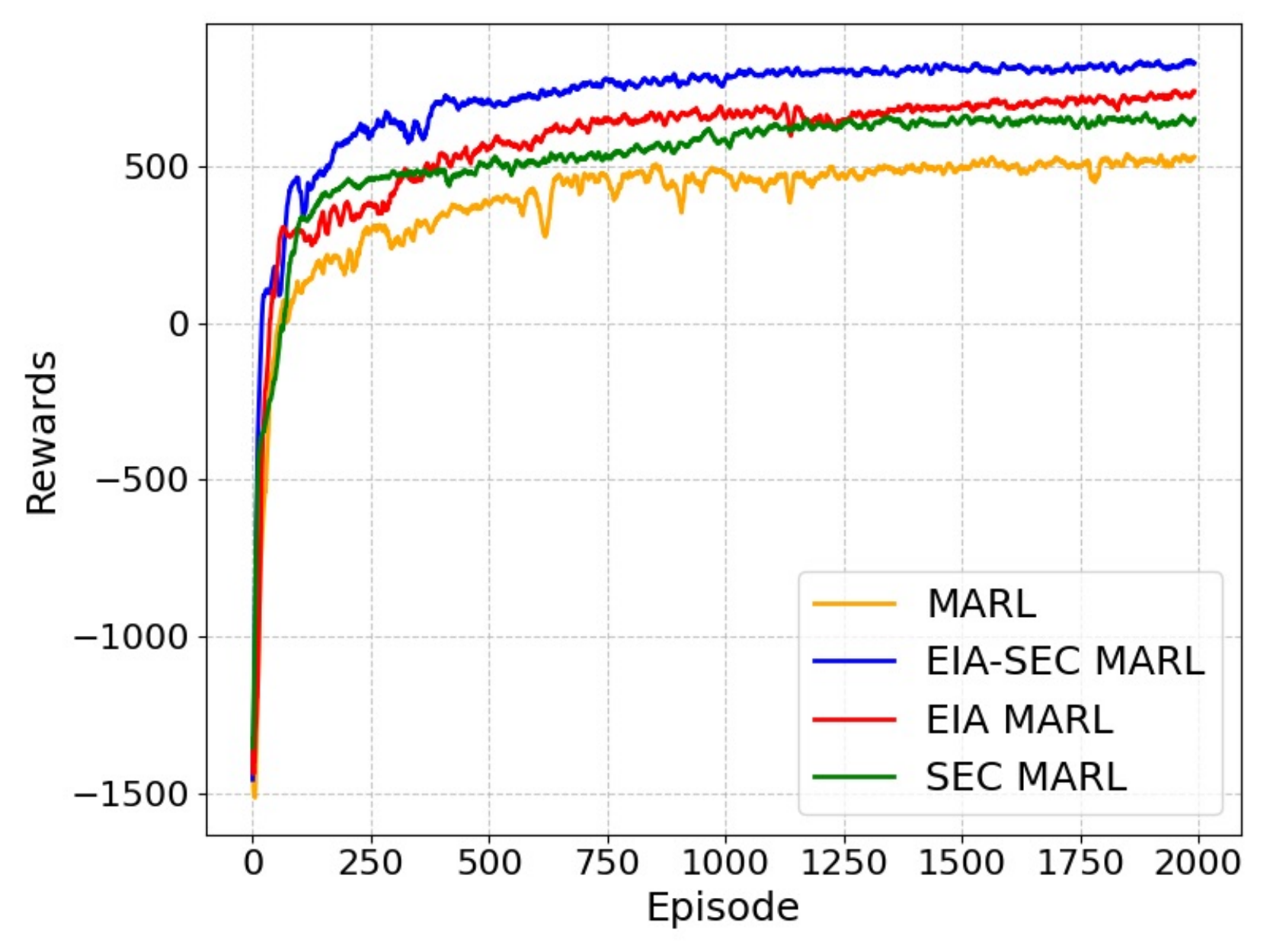}
    \caption{Reward in ablation study.}
    \label{fig: ablation}
    \setlength{\belowcaptionskip}{-10pt}
    \end{subfigure}
            \begin{subfigure}{0.3\textwidth}
        \centering
        \includegraphics[width=\textwidth, keepaspectratio]{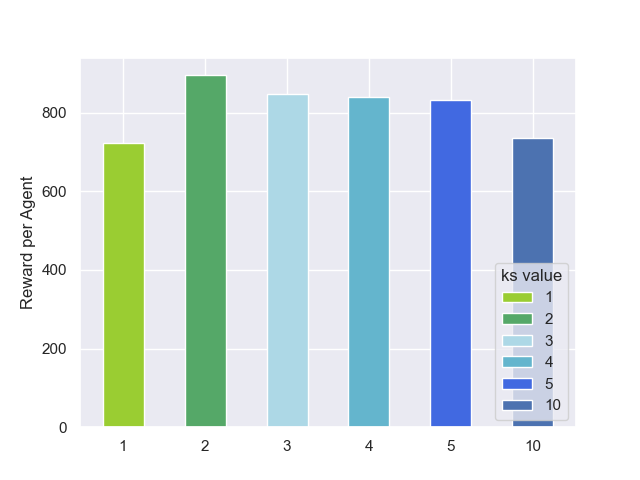}
    \caption{Reward of EIA-SEC with different $k_s$.}
    \label{fig: ks}
    \setlength{\belowcaptionskip}{-10pt}
    \end{subfigure}
    \begin{subfigure}{0.35\textwidth}
        \centering
    \includegraphics[width=\textwidth, keepaspectratio]{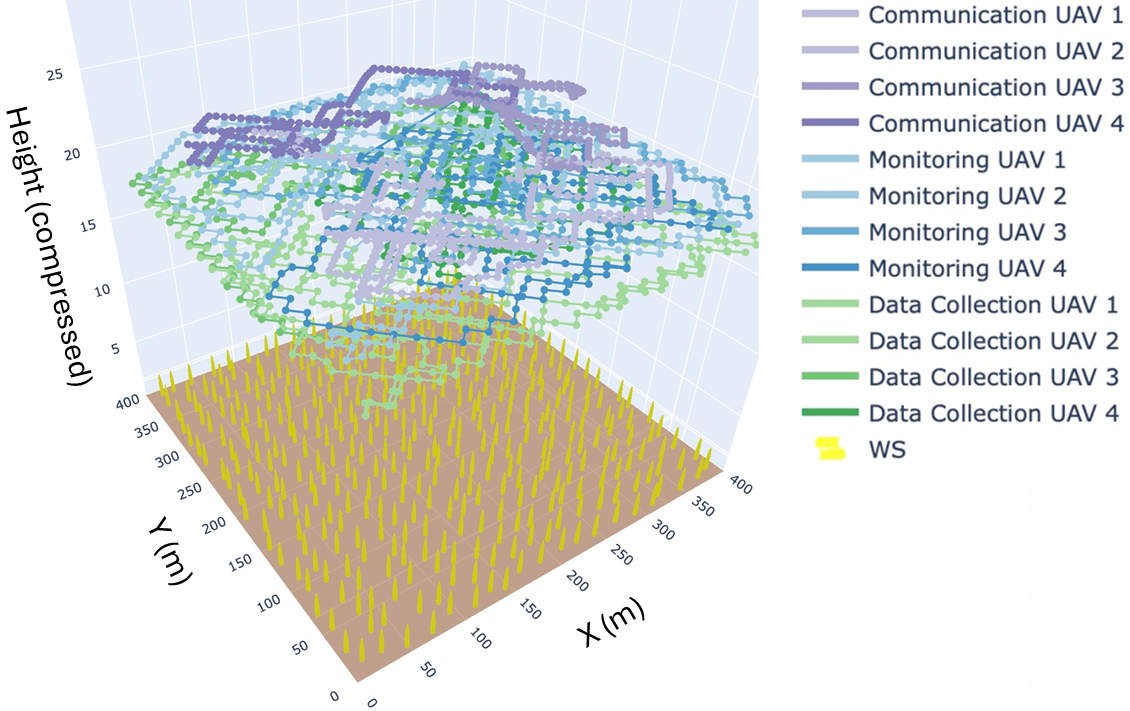}
    \caption{Visualization of UAV trajectories.}
    \label{fig:visualization}
    \setlength{\belowcaptionskip}{-10pt}
    \end{subfigure}
    \caption{Experimental results in the performance, scalability, ablation, and visualization studies.}
    \end{figure*}

We adopt a simulation based on real-world layout\footnote{Referring to the intelligent farm layout and UAV data used in the Zhuozhou Teaching Experimental Farm~\cite{zhuozhou}.}.
Tables~\ref{env_agri} and \ref{hyper_agri} present the environmental settings and hyperparameters, respectively.
We evaluate the EIA-SEC from four perspectives: training performance, scalability and complexity analysis, ablation study, and 3D visualization. 
To enhance generalizability, we train the models on 20 different farmland scenarios with different WSN distributions and test them on a random one.
We define the following performance metrics:
\begin{itemize}
    \item Reward (M1): This metric measures the average reward per agent received during one episode.
    \item Convergence time (M2): This metric measures the number of iterations it takes for the algorithm to converge.
    \item Stability and Generalization (M3): This metric measures the fluctuations after convergence with the dynamic scenarios.
% \end{itemize}
%     In addition, inference time refers to the duration required for agents to make decisions during execution. There are three metrics for evaluating the inference time as follows.
% \begin{itemize}
    \item Scalability (M4): This metric measures how the inference time per timestep varies with the number of UAVs.
    \item Policy Efficiency (M5): This metric measures the inference time of UAVs during tasks.
\end{itemize}

Note that the metrics M1-M3 are measured from the training results, while metrics M4 and M5, as well as UAV trajectories, are obtained from the testing results. We set $n_{\text{c}} = 4$, $n_{\text{d}} = 4$, and $n_{\text{m}} = 4$ except for the scalability study.

\subsection{Training Performance} \label{training_performance}

% In this experiment, we set $n_{\text{c}} = 4$, $n_{\text{d}} = 4$, and $n_{\text{m}} = 4$. 
% Due to the complex and dynamic environment of the scenarios, as well as the limitation of non-RL algorithms, such as heuristics, in handling the coordination among UAVs with different task functions, we only consider comparison experiments with other DRL algorithms.
The non-RL algorithms, e.g., heuristics, fail to handle multi-task control in complex and dynamic environments. Therefore,
we select three state-of-the-art MARL algorithms as baselines: MASAC~\cite{sac}, MATD3~\cite{td3}, and MADDPG~\cite{ddpg}, which are based on the actor-critic framework and can handle continuous action spaces.
% These baseline algorithms are introduced in Section~\ref{related_works}.

\subsubsection{Evaluation of M1} As shown in Fig.~\ref{fig: reward_performance}, the proposed EIA-SEC achieves significantly higher average rewards than all DRL baselines. This improvement can be attributed to the EIA module, which enhances overall performance by learning from elite individuals and preventing the agents from rapidly converging to local optima. In addition, the SEC component mitigates the overestimation of state-action values, thereby improving the reliability and effectiveness of policy selection. MADDPG, MATD3, and MASAC exhibit similar reward levels, lower than EIA-SEC.
As illustrated in Figs.~\ref{fig:data_aoi} and \ref{fig:visual_aoi}, EIA-SEC achieves the best performance in both AoI and VDF, significantly outperforming all other baselines. This observation is consistent with the results shown in Fig.~\ref{fig: reward_performance}, indicating that EIA-SEC can guarantee the timeliness of both data collection and image acquisition.

% \begin{figure}[H]
%     \centering
%     \includegraphics[width=0.38\textwidth, keepaspectratio]{fig/fig-reward.pdf}
%     \caption{Reward performance comparison with other DRL algorithms.}
% \label{fig: reward_performance}
%     \setlength{\belowcaptionskip}{-10pt}  % 调整标题下方间距
% \end{figure}

\subsubsection{Evaluation of M2} As shown in Fig.~\ref{fig: reward_performance}, all approaches converge within 800 episodes, while EIA-SEC yields a higher convergence speed.
% From the convergence perspective, EIA-SEC achieves higher reward value at earlier episode. 
This indicates that the elite selection and learning strategy employed in EIA effectively reduces the agents’ learning and exploration costs, enabling them to escape local optima. Meanwhile, SEC ensures that the critics can learn the state-action value estimations more efficiently, thereby providing more stable and reliable evaluations.

\subsubsection{Evaluation of M3} 
Moreover, the reward curve of EIA-SEC is more stable compared to other baselines.
EIA prevents agents from drifting toward non-optimal learning directions and deviating from the optimal policy, while SEC mitigates overestimation of state-action values, thus preventing agents from making inappropriate action decisions.

Therefore, it can be concluded that EIA-SEC achieves best M1 performance, consistently surpassing all the DRL baselines in reward. For M2, EIA-SEC shows faster convergence than other algorithms, yielding higher training efficiency.
% even delivers better reward performance when evaluated at the same episode. 
Furthermore, its M3 metrics exhibit obvious advantages over DRL baselines, showing higher convergence stability.

% \begin{figure}[H]
%     \centering
%     \includegraphics[width=0.38\textwidth, keepaspectratio]{fig/Data-AoI.pdf}
%     \caption{AoI performance comparison with other DRL algorithms.}
%     \label{fig:data_aoi}
%     \setlength{\belowcaptionskip}{-10pt}  % 调整标题下方间距
% \end{figure}

% \begin{figure}[H]
%     \centering
%     \includegraphics[width=0.38\textwidth, keepaspectratio]{fig/Visual-AoI.pdf}
%     \caption{VDF performance comparison with other DRL algorithms.}
%     \label{fig:visual_aoi}
%     \setlength{\belowcaptionskip}{-10pt}  % 调整标题下方间距
% \end{figure}

\subsection{Scalability and Complexity Analysis} \label{complexity}
In this experiment, we first vary the number of the total UAVs $n_{\text{UAV}}$ from 6 to 24 to analyze the computational complexity. 
Then, we set $n_{\text{UAV}} = 12$, and compare the inference time of proposed EIA-SEC and other baseline algorithms. % introduced in Section~\ref{training_performance}. 

\subsubsection{Evaluation of M4}
% Table~\ref{tab:inference_rewards_agents} 
Figs. \ref{fig: scalable_in} and \ref{fig: scalable_re}
illustrate that although the inference time of the proposed EIA-SEC increases with the number of agents, the average reward per agent remains relatively stable. This is because, as the number of agents grows, the algorithm requires additional computation to analyze inter-UAV positional relationships and trajectory decision-making, which increases the inference overhead. Nevertheless, this increase remains within an acceptable range and is much smaller than the UAVs’ actual task execution time. Therefore, we conclude that EIA-SEC demonstrates good scalability when the number of UAVs increases.

% \begin{table}[htbp]
% \centering
% \caption{Inference Time and Rewards under Different Numbers of Agents.}
% \begin{tabular*}{0.3\textwidth}{@{\extracolsep{\fill}}c|c|c@{}}
% \toprule
% \textbf{Agent} & \textbf{Inference Time (ms)} & \textbf{Rewards} \\
% \midrule
% 6  & 17.64 & 904.09 \\
% 12 & 34.38 & 895.04 \\
% 16 & 44.45 & 864.44 \\
% 20 & 80.81 & 871.08 \\
% 24 & 117.16 & 881.34 \\
% \bottomrule
% \end{tabular*}
% \label{tab:inference_rewards_agents}
% \end{table}

\subsubsection{Evaluation of M5} For inference time, we compare the proposed EIA-SEC with DRL baselines. 
% Table.~\ref{tab: inference_compare} 
Fig.~\ref{fig: scalable_com}
shows that the inference time of EIA-SEC is on the same order of magnitude as that of the other DRLs, with MADDPG performing the best. 
Nevertheless, the performance of MADDPG is much lower than EIA-SEC since it tends to get stuck in local optima, and it is prone to overestimation of state-action values. 
Furthermore, the introduction of the EIA-SEC module does not lead to a significant increase in inference time. 
This is because during the offline decision-making phase, EIA does not introduce additional overhead for elite learning, and the computational cost of elite selection and learning is much smaller than that of completing an episode, thus having no impact on inference time. In addition, although the SEC module introduces extra computational cost for state-action value evaluation, the resulting interactions are far less intensive than the overall system overhead of UAV task decision-making. Moreover, since this process occurs during the offline phase, it does not affect online decision-making. 

% \begin{table}[htbp]
% \centering
% \caption{Inference Time of EIA-SEC and other DRLs.}
% \begin{tabular*}{0.3\textwidth}{@{\extracolsep{\fill}}c|c@{}}
% \toprule
% \textbf{Algorithm} & \textbf{Inference Time (ms)}\\
% \midrule
% EIA-SEC    & 34.38  \\
% MATD3       & 32.21 \\
% MASAC       & 33.26  \\
% MADDPG & 29.27  \\
% \bottomrule
% \end{tabular*}
% \label{tab: inference_compare}
% \end{table}

\subsection{Ablation Study}
% In this experiment, we set $n_{\text{c}} = 4$, $n_{\text{d}} = 4$, and $n_{\text{m}} = 4$.
We evaluate the effectiveness of the proposed EIA and SEC modules. As shown in Fig.~\ref{fig: ablation}, the inclusion of the EIA and SEC modules significantly enhances the reward performance. 
% Without the EIA or SEC module, the reward levels of these models are lower, and convergence is slower, with basic MARL achieving the lowest reward among all configurations. 
This improvement can be attributed to the EIA module, which generates additional experience by selecting and learning from elite individuals. These elite experiences provide valuable guidance to other agents, enabling them to learn more effectively and share strategies across the population. As a result, the agents can reduce the cost of trial-and-error learning, leading to faster convergence, higher reward levels, and greater stability in training. Besides, the SEC module introduces additional critics to help agents estimate state-action values more quickly and efficiently. Through inter-agent negotiation, it also ensures consistency in the evaluation schemes among agents and mitigates value overestimation, thereby improving convergence speed and overall reward performance.
% In Table.~\ref{tab: k-value}, the 
Furthermore, we depict the impacts of the SEC structure on the reward performance in Fig.~\ref{fig: ks}, where
$k_{\text{s}}$-value represents the number of critics in the SEC module. The result shows that a higher $k_{\text{s}}$-value does not necessarily lead to better reward performance for the agents, and 2 critics in the SEC module yield the best performance in the considered scenario. This is because excessive negotiation and learning from other agents’ evaluation strategies can reduce the agents’ exploration ability. Therefore, to achieve better overall performance, the k-value should be maintained within a reasonable range.
In summary, the EIA and SEC modules play a crucial role in accelerating learning while simultaneously improving both efficiency and robustness of the multi-agent collaborative control.

% \begin{figure}[H]
%     \centering
%     \includegraphics[width=0.42\textwidth, keepaspectratio]{fig/ablation.pdf}
%     \caption{Reward comparison in ablation study.}
%     \label{fig: ablation}
%     \setlength{\belowcaptionskip}{-10pt}  % 调整标题下方间距
% \end{figure}

% \begin{table}[htbp]
% \centering
% \caption{Rewards under Different Number Critics of SECs}
% \begin{tabular*}{0.3\textwidth}{@{\extracolsep{\fill}}c|c|c|c@{}}
% \toprule
% \textbf{$k_{\text{s}}$-Value} & \textbf{Rewards} & \textbf{$k_{\text{s}}$-Value} & \textbf{Rewards} \\
% \midrule
% 1  & 721.91 & 4  & 838.38 \\
% \textbf{2}  & \textbf{895.04} & 5  & 832.43 \\
% 3 & 847.91 &  10 & 734.32\\
% \bottomrule
% \end{tabular*}
% \label{tab: k-value}
% \end{table}

\subsection{Visualization}

% We generate 10 different scenarios for 2D UAV trajectory visualization.
Fig.~\ref{fig:visualization} shows an exemplary UAV trajectory visualization\footnote{More visualization results are available at \href{https://anonymous.4open.science/r/Mobisys_smart_agriculture-10EB}{[link to visualization results]}.},
% where $n_{\text{c}} = 4$, $n_{\text{d}} = 4$, and $n_{\text{m}} = 4$. 
which illustrates the collaborative behavior among different types of UAVs.
% which jointly patrol the entire farmland to minimize both AoI and VDF in the intelligent agriculture system. 
Among them, data collection and monitoring UAVs have more regular shapes to jointly patrol the entire farmland to minimize both AoI and VDF in the intelligent agriculture system, while communication UAVs have more irregular routes among other types of UAVs, serving as auxiliary aerial platforms to provide communication services. 

% \begin{figure}[H]
%     \centering
%     \includegraphics[width=0.42\textwidth, keepaspectratio]{fig/Visualization.pdf}
%     \caption{Visualization of UAV trajectories.}
%     \label{fig:visualization}
%     \setlength{\belowcaptionskip}{-10pt}  % 调整标题下方间距
% \end{figure}

\section{Conclusion}
\label{conclusion}
We targeted a multi-UAV smart agricultural system, where the UAVs for data collection, image acquisition, and communication tasks work collaboratively.
We developed a MARL framework, EIA-SEC, that enhances both decision-making stability and generalization capability for multi-agent systems operating in dynamic and complex environments. 
% By introducing the EIA mechanism, agents periodically reference the behavior of high-performing individuals, effectively reducing trial-and-error exploration costs while accelerating convergence toward high-quality policies. Meanwhile, agents gradually diminish their dependence on elite individuals, thereby preserving exploration ability and avoiding premature convergence to suboptimal solutions.
% To further improve the objectivity and reliability of value estimation, we proposed the SEC mechanism, in which each agent’s local critic collaborates with a shared ensemble of critics from agents performing the same task. This design mitigates the risk of Q-value overestimation and stabilizes the learning process, providing more robust value guidance throughout training.
% Our proposed EIA-SEC framework significantly enhances learning efficiency, stability, and scalability in complex collaborative environments. 
Within the context of intelligent agricultural UAV systems, our approach demonstrates improved convergence behavior and superior performance in coordinated sensing and monitoring tasks. 
Overall, the results indicate that elite-guided policy learning and shared critic ensembles offer a promising direction for advancing MARL in real-world applications.

%Bibliography
\bibliographystyle{unsrt}  
\bibliography{references}

\end{document}